\documentclass[sigconf]{acmart}

\usepackage{comment}
\usepackage{titlesec}
\usepackage{graphicx}
\usepackage{subcaption}
\usepackage{algorithm}

\usepackage{algpseudocode}
\usepackage{booktabs}
\usepackage{adjustbox}
\usepackage{xcolor}
\usepackage[utf8]{inputenc}
\usepackage{CJK}

\usepackage{amssymb}
\usepackage{amsthm}
\usepackage{amsmath}
\usepackage{multirow}
\usepackage{makecell}

\newtheorem{theorem}{Theorem}
\newtheorem{definition}{Definition}
\newtheorem{lemma}{Lemma}
\newtheorem{corollary}{Corollary}

\usepackage[online]{optional}

\newcommand{\phead}[1]
{
\noindent \textbf{#1}
}

\newcommand{\shenr}[1]{
{\color{blue}{#1}}
}

\newcommand{\luICDE}[1]
{
{\color{orange}{#1}}
}

\copyrightyear{2025}
\acmYear{2025}
\setcopyright{acmlicensed}\acmConference[CIKM '25]{Proceedings of the 34th ACM International Conference on Information and Knowledge Management}{November 10--14, 2025}{Seoul, Republic of Korea}
\acmBooktitle{Proceedings of the 34th ACM International Conference on Information and Knowledge Management (CIKM '25), November 10--14, 2025, Seoul, Republic of Korea}
\acmDOI{10.1145/3746252.3760994}
\acmISBN{979-8-4007-2040-6/2025/11}

\begin{document}
\begin{CJK}{UTF8}{bkai}
\makeatletter

\renewcommand\subsubsection{\@secnumfont}{\bfseries}%
\renewcommand\subsubsection{\@startsection{subsubsection}{3}
  \z@{.5\linespacing\@plus.7\linespacing}{-.5em}%
  {\normalfont\bfseries}}
  
  \makeatother
\title{Enhancing Contrastive Link Prediction With Edge Balancing Augmentation}

\author{Chen-Hao Chang}
\email{s111062579@m111.nthu.edu.tw}
\affiliation{%
  \institution{National Tsing Hua University}
  \department{Department of Computer Science}
  \city{Hsinchu}
  \country{Taiwan}
}

\author{Hui-Ju Hung}
\email{hjhung@ncu.edu.tw}
\affiliation{%
  \institution{National Central University}
  \department{Department of Computer Science and Information Engineering}
  \city{Taoyuan}
  \country{Taiwan}
}

\author{Chia-Hsun Lu}
\email{chlu@m109.nthu.edu.tw}
\affiliation{%
  \institution{National Tsing Hua University}
  \department{Department of Computer Science}
  \city{Hsinchu}
  \country{Taiwan}
}

\author{Chih-Ya Shen}
\email{chihya@cs.nthu.edu.tw}
\affiliation{%
  \institution{National Tsing Hua University}
  \department{Department of Computer Science}
  \city{Hsinchu}
  \country{Taiwan}
}



\opt{full}{ 
\begin{abstract}
Link prediction is one of the most fundamental tasks in graph mining, which motivates the recent studies of leveraging contrastive learning to enhance the performance. However, we observe two major weaknesses of these studies: i) the lack of theoretical analysis for contrastive learning on link prediction, and ii) inadequate consideration of node degrees in contrastive learning. To address the above weaknesses, we provide the first formal theoretical analysis for contrastive learning on link prediction, where our analysis results can generalize to the autoencoder-based link prediction models with contrastive learning. Motivated by our analysis results, we propose a new graph augmentation approach, \emph{Edge Balancing Augmentation (EBA)}, which adjusts the node degrees in the graph as the augmentation. We then propose a new approach, named \emph{Contrastive Link Prediction with Edge Balancing Augmentation} (\texttt{CoEBA}), that integrates the proposed EBA and the proposed new contrastive losses to improve the model performance. We conduct experiments on 9 benchmark datasets, including the widely-used OGB dataset, and also evaluate our approach under the challenging \emph{HeaRT setting}. The results demonstrate that our proposed \texttt{CoEBA} significantly outperforms the other state-of-the-art link prediction models.
\end{abstract}
}
\opt{short}{
\begin{abstract}
Link prediction is one of the most fundamental tasks in graph mining, which motivates the recent studies of leveraging contrastive learning to enhance the performance. However, we observe two major weaknesses of these studies: i) the lack of theoretical analysis for contrastive learning on link prediction, and ii) inadequate consideration of node degrees in contrastive learning. To address the above weaknesses, we provide the first formal theoretical analysis for contrastive learning on link prediction, where our analysis results can generalize to the autoencoder-based link prediction models with contrastive learning. Motivated by our analysis results, we propose a new graph augmentation approach, \emph{Edge Balancing Augmentation (EBA)}, which adjusts the node degrees in the graph as the augmentation. We then propose a new approach, named \emph{Contrastive Link Prediction with Edge Balancing Augmentation} (\texttt{CoEBA}), that integrates the proposed EBA and the proposed new contrastive losses to improve the model performance. We conduct experiments on 8 benchmark datasets. The results demonstrate that our proposed \texttt{CoEBA} significantly outperforms the other state-of-the-art link prediction models.
\end{abstract}
}
\opt{online}{
\begin{abstract}
Link prediction is one of the most fundamental tasks in graph mining, which motivates the recent studies of leveraging contrastive learning to enhance the performance. However, we observe two major weaknesses of these studies: i) the lack of theoretical analysis for contrastive learning on link prediction, and ii) inadequate consideration of node degrees in contrastive learning. To address the above weaknesses, we provide the first formal theoretical analysis for contrastive learning on link prediction, where our analysis results can generalize to the autoencoder-based link prediction models with contrastive learning. Motivated by our analysis results, we propose a new graph augmentation approach, \emph{Edge Balancing Augmentation (EBA)}, which adjusts the node degrees in the graph as the augmentation. We then propose a new approach, named \emph{Contrastive Link Prediction with Edge Balancing Augmentation} (\texttt{CoEBA}), that integrates the proposed EBA and the proposed new contrastive losses to improve the model performance. We conduct experiments on 8 benchmark datasets. The results demonstrate that our proposed \texttt{CoEBA} significantly outperforms the other state-of-the-art link prediction models.
\end{abstract}
}



\maketitle

\section{Introduction}\label{sec:intro}

Link prediction, one of the most fundamental tasks in graph mining and machine learning on graphs, aims at predicting the
existence of potential links between nodes on a graph.
Effective link prediction helps identify potential relationships and enhance connectivity within a network. Applications of link prediction include recommendation, e-commerce, friend recommendation in social networks, and much more~\cite{Fan2022,Wang2023recommend,Kundu_2023_CVPR}.


To construct an accurate link prediction model, abundant labeled data are required to ensure the performance. However, in real-world scenarios, labeled data are usually scarce and expensive to obtain.
Moreover, real-world graphs often contain numerous missing edges, which can significantly distort key graph properties~\cite{haj2022inferring,kossinets2006effects}. Consequently, the graph data tend to be noisy, as the presence or absence of an edge in the observed graph does not necessarily reflect the true underlying relationship between two entities. 
To address this challenge, recent studies \cite{Li2024toward,liang2023knowledge,zhang2023line} employ contrastive learning to effectively improve the performance of machine learning on graphs. The fundamental concept of contrastive learning involves developing robust embeddings by contrasting positive samples against negative samples \cite{you2020}. For contrastive learning on graph-related tasks, graph augmentation is usually performed to generate positive and negative examples to facilitate the learning of robust graph neural networks. 
Although existing contrastive learning approaches achieve promising performance in graph-related tasks, there still exist some issues, particularly when contrastive learning is adopted for link prediction, as identified below.

\phead{Weakness W1. Lack of theoretical analysis for contrastive learning on link prediction.}
Contrastive learning shows outstanding performance in node classification tasks; however, employing contrastive learning for link prediction is still in its early stage. Recent studies on the theoretical aspects of contrastive learning primarily focus on node classification tasks, and the inherent differences between node classification and link prediction often render these analyses inapplicable to the latter. 
However, the theoretical exploration of how contrastive learning functions, specifically within the domain of link prediction, remains unexplored.
 
\phead{Weakness W2. Inadequate consideration of node degrees in graph augmentation.} 
Graph augmentation is a crucial component in graph contrastive learning, as improper augmentation may harm the model performance. Recent research has incorporated semantic relationships into the design of graph augmentation strategies~\cite{Chen_Kou_2023,Li2024toward,yin2022}, aiming to leverage the inherent information of the graph to create suitable augmentations. 
On the other hand, as suggested by our theoretical analysis in this paper, the node degrees in the graph are closely related to the link prediction performance. However, existing approaches fail to consider the node degrees in their graph augmentation strategies. This oversight results in augmented graphs that are either sparser or not significantly different from the original, leading to significantly inferior performance.

To address the above weaknesses, we first provide a formal theoretical analysis for contrastive learning on link prediction (to address weakness W1). Our theoretical analysis identifies the important factors that are closely related to the error rate of the link prediction performance. Please note that our theoretical analysis is very general, which holds for all the common autoencoder-based link prediction models with contrastive learning.

Our analysis results also motivate our design of the proposed approach, \emph{Contrastive Link Prediction with Edge Balancing Augmentation} (\texttt{CoEBA}), which includes a new graph augmentation strategy, named \emph{Edge Balancing Augmentation} (EBA). The novelty of EBA lies on the adjustments of the degrees as the augmentation in the graph, which allows the contrastive learning to effectively improve its performance (to address weakness W2). 
The proposed EBA significantly improves the concentration of node embeddings in the latent space, thereby enhancing the generalization of contrastive learning. Since the proposed EBA is supported by our theoretical analysis, it can also benefit the autoencoder-based link prediction models, such as \texttt{GAE}~\cite{kipf2016}, \texttt{GNAE}~\cite{ahn2021}, \texttt{VGNAE}~\cite{ahn2021}, and many more, as will be illustrated in the experiments.

\opt{full}{
In addition, we also propose the \emph{neighbor-concentrated contrastive losses} to enrich node embeddings with additional structural information. By considering neighborhood structure as positive samples, we ensure that embeddings of connected nodes are drawn closer together, effectively capturing the underlying graph topology.
We perform extensive experiments to evaluate our proposed \texttt{CoEBA} on 9 benchmark datasets, including the widely-used OGB dataset. Furthermore, we evaluate our method under the challenging \emph{HeaRT setting}~\cite{Li2023evaluate} to assess its robustness in difficult scenarios. 
}
\opt{short}{
In addition, we also propose the \emph{neighbor-concentrated contrastive losses} to enrich node embeddings with additional structural information. By considering neighborhood structure as positive samples, we ensure that embeddings of connected nodes are drawn closer together, effectively capturing the underlying graph topology. We perform extensive experiments to evaluate our proposed \texttt{CoEBA} on 8 benchmark datasets.
}
\opt{online}{
In addition, we also propose the \emph{neighbor-concentrated contrastive losses} to enrich node embeddings with additional structural information. By considering neighborhood structure as positive samples, we ensure that embeddings of connected nodes are drawn closer together, effectively capturing the underlying graph topology. We perform extensive experiments to evaluate our proposed \texttt{CoEBA} on 8 benchmark datasets.
}
The results show that our approach outperforms other state-of-the-art models. 

The contributions are summarized as follows.
\begin{itemize}
  \item To the best of our knowledge, we are the first to provide the theoretical analysis of the performance on link prediction with contrastive learning. The analysis results generalize to the common autoencoder-based link prediction models with contrastive learning.
  \item Motivated by the analysis results, we propose a new graph augmentation method,  called \emph{Edge Balancing Augmentation (EBA)}. EBA is the first to explicitly adjust the node degrees to enhance the link prediction performance, and EBA can serve as a plug-and-play module to enhance other autoencoder-based link prediction models.
  
  \item  We propose the new \texttt{CoEBA} approach that integrates the \emph{EBA} and  \emph{neighbor-concentrated contrastive losses} to enrich node embeddings with structural information. 
  
  \item Extensive experimental results demonstrate that our proposed \texttt{CoEBA} significantly outperforms other state-of-the-art link prediction models.
\end{itemize}

The paper is organized as follows. Sec.~\ref{sec:related} discusses the related work. Sec.~\ref{sec:preliminary} formulates the problem and provides the theoretical analysis. Sec.~\ref{sec:algo} describes the design of our proposed approach. Sec.~\ref{sec:exp} presents the experimental results. Sec.~\ref{sec:conclusion} concludes this paper.

\section{Related Work}
\label{sec:related}

\phead{Link prediction.}
Recent GNN-based link prediction methods can be roughly divided into two categories, \emph{node-embedding-based} and \emph{subgraph-based}. 
\textit{Node-embedding-based methods} employ autoencoders to learn embeddings from local neighborhoods and aggregate pairwise node embeddings to predict links~\cite{ahn2021,guo2022,tan2022,yun2022,wang2024,pmlr-v70-gilmer17a}. For example, \texttt{VGNAE}~\cite{ahn2021} adopts L2-normalization to improve embedding for isolated nodes and introduces variational inference to learn a distribution which represents node embeddings; 
\texttt{S2GAE}~\cite{tan2022} improves the generalizability of graph autoencoders (GAEs) with a novel graph masking strategy and employs a tailored cross-correlation decoder to capture cross-correlation of their end nodes from multi-granularity embeddings. \texttt{Neo-GNNs}~\cite{yun2022} enhances GNNs by leveraging structural features from adjacency matrices to better estimate neighborhood overlaps for link prediction. \texttt{NCNC}~\cite{wang2024} leverages structural features to enhance the pooling process. 

On the other hand, the core idea of \textit{subgraph-based methods} is to extract the $h$-hop neighbors of the target links as their subgraphs to represent the local neighborhood structure between nodes. \texttt{SEAL}~\cite{zhang2018} employs GNNs to learn heuristics from local subgraphs by utilizing a $\gamma$-decaying theory. \texttt{BUDDY}~\cite{chamberlain2023} enhances link prediction by integrating subgraph sketches and precomputed features to address the inefficiencies of traditional subgraph GNNs and improve expressiveness and scalability. 
\texttt{ML-Link}~\cite{zangari2024link} augments multilayer GNNs with node-pair structural features learned from both within-layer and across-layer overlapping neighborhoods to improve link prediction. 
\texttt{PULL}~\cite{kim2025accurate} treats link prediction as PU learning to recover latent links from unobserved node pairs. 

Despite recent research shows that subgraph-based methods are more expressive, subgraph-based methods usually suffer from intensive computational costs. On the other hand, node-embedding-based approaches are usually more computationally efficient while preserving comparable accuracy.
In contrast, the proposed \texttt{CoEBA} model integrates augmented graphs generated based on our theoretical analysis to enrich the training data, which helps in capturing the semantic relationships and structural similarities between nodes more effectively and results in a more generalizable model.

\phead{Contrastive learning.}
Contrastive learning is particularly effective in unsupervised and semi-supervised settings by employing data augmentations to create positive and negative pairs, optimizing a contrastive loss that promotes invariance in the learned embeddings. Inspired by advances in computer vision~\cite{chen2020}, recent works extend this to graphs-related tasks. GraphCL~\cite{you2020} uses random augmentations (e.g., node dropping, edge flipping), GCA~\cite{zhu2021} leverages centrality-guided edge perturbations, and AutoGCL~\cite{yin2022} introduces learnable view generators for adaptive and structure-preserving augmentation.


Recently, several studies have analyzed the theoretical properties of contrastive learning. Wang et al.~\cite{pmlr-v119-wang20k} demonstrate that uniformity and alignment are two crucial properties in contrastive learning for computer vision tasks. Huang et al.~\cite{huang2023towards} and Wang et al.~\cite{Wang2022} analyze the factors that improve the generalization ability for node classification. Tian et al.~\cite{NEURIPS2022_Tian, tian2023understanding} formulate contrastive learning as a min-max problem and further analyze the role of nonlinearity in contrastive learning under the context of image classification. 
Tan et al.~\cite{tan2024contrastive} show that contrastive learning is equivalent to performing spectral clustering on a similarity graph. 

In this paper, different from previous research, we explore the integration of contrastive learning with node-embedding-based link prediction approaches. We propose the augmentation strategy based on node degrees to enhance the generalization ability, aiming to combine the rapid computation of node-based approaches with the ability to learn diverse information through a carefully designed contrastive learning approach.

\section{Problem Formulation \&  Theoretical Analysis}
\label{sec:preliminary}

We first formulate the link prediction task and then present our theoretical analysis. To the best of our knowledge, our theoretical analysis is the first one of contrastive learning on link prediction. Later in Sec.~\ref{sec:algo}, we propose \texttt{CoEBA} based on our theoretical analysis to effectively enhance the performance.

\subsection{Problem Formulation}
Given an undirected graph $G=(V,E,X)$, where $V$ denotes the node set $V=\{v_1,v_2, ...,v_N\}$ with size $N$, and $E \subseteq V \times V$ denotes the edge set. We denote the feature matrix as $X\in\mathbb{R}^{N\times D_f}$, where $x_i\in X$ with dimension $D_f$ is the feature of node $v_i$. Let $(v_i,v_j)\in E$ denote that the nodes $v_i$ and $v_j$ share an edge, and $A\in \{0,1\}^{N\times N}$ represents the adjacency matrix of $G$, where $a_{ij}=1$ if and only if $(v_i,v_j)\in E$. 
We denote the neighbor set of node $v_i$ in graph $G$ as $N_G(v_i)=\{v_j\in V|(v_i,v_j)\in E\}$.


In this work, following prior research \cite{zhao2022, yang2023}, we define the link prediction problem as follows. Given an incomplete edge set $E_{incomplete}$ of a graph $G=(V, E, X)$, our goal is to identify a predictive function $F$ that takes the nodes $V$, the edges $E$, and the node features $X$ as inputs to predict the ideal edge set $E_{ideal}$. The objective is to ensure that the predicted edge set closely approximates the ideal edge set.

\subsection{Review of Previous Theoretical Foundations}
\label{sec:theoretical}
We start our analysis by first reviewing some important theoretical results previously discussed for the node classification task. We then extend the theoretical derivation and discussion to link prediction in Sec.~\ref{subsec:our_theoretical_part}. Please note that as link prediction is more complex than node classification, i.e., considering node pairs instead of a single node in the node classification task, our analysis in Sec.~\ref{subsec:our_theoretical_part} is the first theoretical derivation to connect the minimum node degrees to the link prediction performance.

\phead{Basic concept for contrastive learning in \emph{node classification}.}
As discussed in previous research~\cite{huang2023towards,Wang2022}, contrastive learning pushes positive samples together and pulls negative samples away. In the meanwhile, the InfoNCE loss can also be decomposed as \emph{alignment} and \emph{divergence} terms, which represent the alignment of positive samples and divergence of the augmented data, respectively.

We are given a set of nodes, where each node belongs to at least one of the $P$ latent clusters, \{$C_1$, $C_2$, ..., $C_P$\}.
We denote the augmentation set as $\mathcal{T}$, and the set of potential positive samples generated from a graph is denoted as $\mathcal{T}(G)$. The objective of graph contrastive learning is to train a graph neural network encoder such that the positive pairs are closely aligned in the learned latent (embedding) space while negative pairs are pushed apart.

To measure the concentration of the embeddings, the concept of $(\alpha, \gamma, \hat{d})$-augmentation is introduced~\cite{Wang2022} for node classification.

\begin{definition}\label{def:agd-aug}
\textbf{$(\alpha, \gamma, \hat{d})$-augmentation~\cite{Wang2022}}. An augmentation set $\mathcal{T}$ is an $(\alpha, \gamma, \hat{d})$-augmentation if for each cluster $C_p$, there exists a subset $C^0_p \subset C_p$ such that the following two conditions hold:
\begin{enumerate}
    \item $\mathbb{P}[v_i \in C^0_p] \geq \alpha \mathbb{P}[v_i \in C_p]$, where $\alpha \in (0,1]$,
    \item $\sup_{v_i,v_j \in C^0_p} d_\mathcal{T}(v_i, v_j) \leq \gamma \left(\frac{B}{\hat{d}^{p}_{\min}}\right)^{\frac{1}{2}}$, where $\gamma \in (0,1]$.
\end{enumerate}
Here, $\hat{d}^{p}_{\min} = \min_{v_i \in C^0_p, \hat{g}_i \in \mathcal{T}(g_i)} \hat{d}_i$, $\hat{d}_i$ is the node degree of node $v_i$, and $B$ is the feature dimension. $\mathbb{P}[v_i]\in C^0_p$ indicates the probability of node $v_i$ belonging to the cluster $C_p$, and $\sup$ is the supremum of a set.
\end{definition}

\begin{figure}[t]
    \centering
    \begin{subfigure}[t]{0.47\columnwidth}
        \centering
        \includegraphics[width=\linewidth]{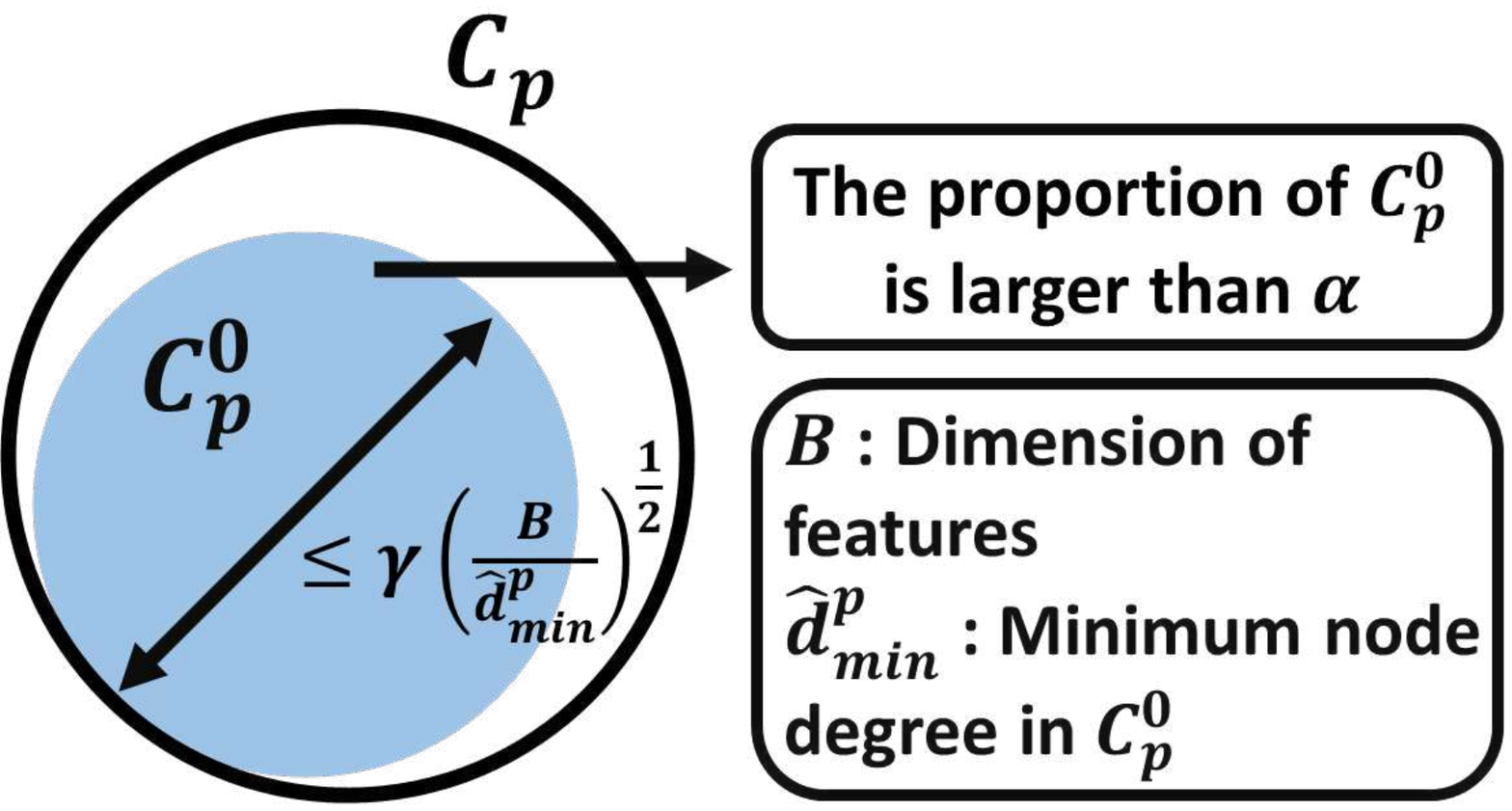}
        \caption{}
        \label{fig:def1}
    \end{subfigure}
    \hfill
    \begin{subfigure}[t]{0.47\columnwidth}
        \centering
        \includegraphics[width=\linewidth]{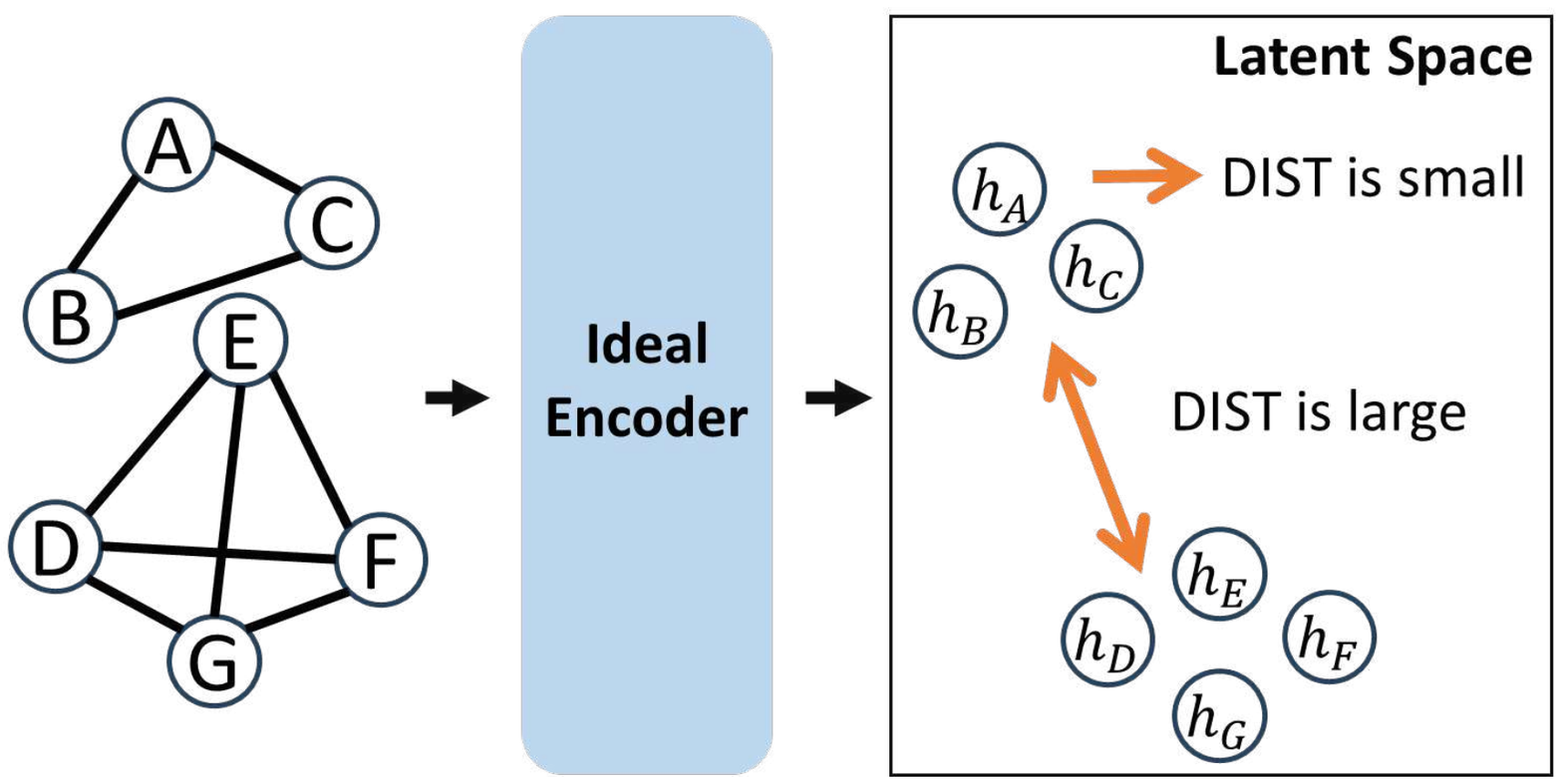}
        \caption{}
        \label{fig:def2}
    \end{subfigure}
    \caption{Illustrations of Def.~\ref{def:agd-aug} and Def.~\ref{def2}. (a) Augmentation Illustration:
        The $(\alpha, \gamma, \hat{d})$-augmentation, where the subset $C^0_p$ occupies at least an $\alpha$ proportion of cluster $C_p$ and forms a ball with diameter at most $\gamma \left(\frac{B}{\hat{d}^{p}_{\min}}\right)^{\frac{1}{2}}$. (b) Ideal Link Prediction: 
         An illustration of the ideal link prediction in the latent space.}
    
    \label{fig:definition}
\end{figure}

The $(\alpha,\gamma, \hat{d})$-augmentation defined in Def.~\ref{def:agd-aug} actually describes the clustering behavior of similar samples in the latent space. The main idea is that the trained embeddings will form clusters in the latent space. Each cluster contains a subset $C^0_p$ that serves as its \emph{main part}, comprising at least an $\alpha$ proportion of the cluster $C_p$. The distance between the data points within $C^0_p$ is smaller than or equal to $\gamma(\frac{B}{\hat{d}^{p}_{min}})^{\frac{1}{2}}$. In other words, a larger $\alpha$ and a smaller $\gamma(\frac{B}{\hat{d}^{p}_{min}})^{\frac{1}{2}}$ result in more concentrated embeddings. 
This concept is illustrated in Fig.~\ref{fig:def1}, which visualizes $C^0_p$ as a concentrated core region within the cluster $C_p$ under the $(\alpha,\gamma, \hat{d})$-augmentation.

\vspace{+5pt}
Wang et al.~\cite{Wang2022} further provide a theoretical guarantee that the encoder trained using contrastive learning has an upper bound on the error rate of the downstream node classification task. Here, the error rate is the probability of incorrectly assigning a sample to a cluster to which it does not belong. This upper bound is determined by the parameter $\alpha$ and the alignment loss. In other words, a better alignment enhances the generalization capability of the downstream node classification tasks. Additionally, achieving a more concentrated augmentation, i.e., with a larger $\alpha$, also contributes to improved generalization. Formally, Theorem~\ref{thm:downstream_error} states the theoretical guarantee, and the detailed proof can be found in~\cite{Wang2022}.  



Specifically, we first summarize important terms~\cite{Wang2022}. Let the closed ego network of $v_i$ on $G$ be $N_G[v_i]=N_G(v_i)\cup \{v_i\}$, and consider a cluster indicator $F_f$ that assigns each node to the nearest community center based on its averaged augmented embeddings. 
Let $S_{\epsilon} = \{ v_i \in \bigcup_{p=1}^{P} C_p | \forall G_{i}^{1}, G_{i}^{2} \in \mathcal{T}(N_G[v_i])\wedge \left\| f(G_{i}^{1}) - f(G_{i}^{2}) \right\| \leq \epsilon \}$ be the set of nodes with $\epsilon$-close embeddings among the graph augmentations, where $\epsilon$-close embeddings are those with distance from each other less than or equal to a small positive value $\epsilon$. Here, $\mathcal{T}$ is the augmentation set and $\left\| \cdot \right\|$ is the distance measurement function. We also denote $R_\epsilon = \mathbb{P}[\overline{S_\epsilon}]$, where $\overline{S_\epsilon}$ is the complement set of $S_\epsilon$.

\begin{theorem}
\label{thm:downstream_error}
Given a cluster indicator $F_{f}$, which indicates whether an encoder can distinguish the embedding well. The downstream error rate of $F_{f}$ is bounded by $(1-\alpha)\ + R_\epsilon$. This indicates that they are considered similar within the threshold $\epsilon$. Furthermore, the term $R_\epsilon$ is bounded by $R_{\epsilon} \leq \frac{\left[ C(N-1, m) \right]^2}{\epsilon} \mathbb{E}_{v_i} \mathbb{E}_{\hat{G_i^1}, \hat{G_i^2} \in \mathcal{T}(G_i)} \left\| f(\hat{G_i^1}) - f(\hat{G_i^2}) \right\|$, where $N$ is the number of nodes, the graph augmentations are uniformly sampled with $m$ augmented edges, and $C$ is the combination, with the augmentations satisfying the definition of $(\alpha,\gamma,\hat{d})$-augmentation \cite{Wang2022}. 
\end{theorem}

To begin with, this theorem is established under the assumption of uniform sampling, a standard premise frequently adopted in theoretical analyses of contrastive learning~\cite{Wang2022,huang2023towards}.
Based on Theorem~\ref{thm:downstream_error} and Def.~\ref{def:agd-aug}, a larger $\alpha$ and a smaller $\gamma \left( \frac{B}{\hat{d}^p_{\min}} \right)^{\frac{1}{2}}$  lead to more concentrated embeddings. Moreover, $R_\epsilon$ can be viewed as the proportion of $\epsilon$-close embeddings. A higher $R_\epsilon$ indicates a more robust model.

\phead{Implications and limitations of Theorem~\ref{thm:downstream_error}.} Theorem~\ref{thm:downstream_error} states that the error rate of the learned encoder to classify a node into the correct cluster (i.e., the class for the node classification problem) can be upper-bounded by $(1-\alpha) + R_\epsilon$. We can thus optimize the bound by designing an augmentation of a sharper concentration to obtain a better performance in the downstream task.

The major limitation of the results in Theorem~\ref{thm:downstream_error} is that it justifies the bounds for \textit{only} node classification tasks. However, it is unclear how this theoretical result can be extended to link prediction. More specifically, for the link prediction problem studied in this paper, our goal is to train an encoder to obtain discriminant node embedding for similarity measurement and predict the link existence based on the learned node embedding. Therefore, in this paper, we provide the first theoretical analysis of the upper bound on the dissimilarity in node embeddings from different clusters \textit{for the link prediction task}. In addition, based on the theoretical analysis, we design a new graph augmentation approach which significantly boosts the performance of link prediction. Please note that our analysis results hold for all the common autoencoder-based models for the link prediction task. 


\subsection{Theoretical Analysis for Contrastive Learning on Link Prediction}\label{subsec:our_theoretical_part}
As previously discussed, in node classification, nodes belonging to the same latent class tend to cluster together due to the influence of contrastive learning. In the context of link prediction, it is anticipated that node pairs which share the same or potential link relationships will also be clustered together (we name this the \emph{Potential-Link Concentration (PLCon)} property). More specifically, the PLCon property states that the nodes within the same latent cluster are more likely to form links than nodes from different latent clusters. Such clustering would enhance the ability of downstream encoders to predict the links. Therefore, we provide the theoretical justification that link prediction, particularly when employing autoencoder-based methods, exhibits the Potential-Link Concentration (PLCon) property. 

\phead{Cluster property of node embedding for link prediction.}
Our first step is to analyze whether the embeddings trained by link prediction models, particularly those based on autoencoders, present the PLCon property. Once the embeddings trained by the contrastive link prediction model exhibit the PLCon property, we can consider the ideal encoder would cluster nodes with existing connections and separate those without, thereby forming distinct clusters. Similar to Theorem~\ref{thm:downstream_error}, we derive an upper bound on the dissimilarity of embeddings from different clusters after the encoder training. By designing the augmentation strategies based on this analysis, we could effectively reduce the encoder's error rate, thereby decreasing the probability of incorrectly clustering nodes with different connectivity properties and benefitting the link prediction problem. We begin by introducing the widely accepted concept of ideal link prediction~\cite{shiao2023link}.

\begin{definition}\label{def2}
\textbf{Ideal link prediction~\cite{shiao2023link}.} 
For some nodes $u, v, w \in V$, let $(u, v) \in E$ and $(u, w) \not\in E$. Then, an ideal encoder for link prediction would have $\text{DIST}(h_u, h_v ) < \text{DIST}(h_u, h_w)$ for some distance function DIST.
\end{definition}

Fig.~\ref{fig:def2} illustrate the concept of Def.~\ref{def2}. 
This definition specifies that for an ideal encoder in a link prediction task, the generated embeddings should ensure that nodes with links have smaller distances between them than nodes without links. Building on this concept, we analyze the clustering properties in link prediction. Since this work aims to tackle the link prediction with contrastive learning, we begin our analysis with the reconstruction loss and contrastive loss.

For the reconstruction loss, we adopt the widely used formulation~\cite{kipf2016,ahn2021,yang2023}, defined as:
\begin{align}
\mathcal{L}_{\text{rec}} = -\mathbb{E}_{q(Z|X,A)}[\log p(A|Z)] + \operatorname{KL}(q(Z|X,A) \| p(Z)), \label{eq_rec}
\end{align}
where $Z$ is the latent embedding, $X$ and $A$ are the input feature and adjacency matrices, respectively. The term $q(Z|X, A)$ corresponds to the inference model (encoder), while $p(A|Z)$ represents the generative model (decoder). The decoder typically applies a sigmoid function to the dot product of embeddings $z_i, z_j \in Z$, i.e., $\sigma(z_i^Tz_j)$. The term $KL(\cdot)$ denotes the Kullback-Leibler (KL)-divergence term. 

For the contrastive loss, we incorporate the commonly used contrastive loss form~\cite{you2020}, defined as:
\begin{align}
\mathcal{L}_{\text{contrastive}} = -\log \frac{\exp(\text{sim}(z_i, z^{\prime}_i) / \tau)}{\sum_{j=1}^{N} \exp(\text{sim}(z_i, z^{\prime}_j) / \tau)}, \label{eq_contrasive}
\end{align}
where $\text{sim}(z, z^{\prime})$ represents the similarity measure between embeddings, typically the cosine similarity, and $\tau$ is a temperature parameter.

\begin{lemma} \textbf{Intra-cluster concentration.}
\label{lemma:cluster}
For a graph autoencoder trained with reconstruction and contrastive losses, the learned node embedding of the connected nodes would be closer in the latent space compared to those of unconnected nodes, such that their pairwise distances are constrained by a threshold $\delta$ for nodes $v_i$ and $v_j$, and $\delta$ is not infinity.
\end{lemma}

\begin{proof}
\opt{short}{Detailed proof is in the online full version~\cite{rep_material}.}
\opt{online}{
With the reconstruction loss in Eq. \ref{eq_rec}, the model maximizes $\mathbb{E}_{q(Z|X,A)}[\log p(A|Z)]$ while minimizing $\operatorname{KL}(q(Z|X,A) \| p(Z))$. Since we predict $A_{pred}$ by calculating the inner product of $Z$, $\sigma(z_i^Tz_j)$, therefore, the model maximizes $z_i^Tz_j$ when $A_{ij}=1$ and minimizes $z_i^Tz_j$ when $A_{ij}=0$.

As for the contrastive loss in Eq. \ref{eq_contrasive}, $\text{sim}(z, z')$ represents the similarity measure between the embeddings. 
Maximizing the similarity of a pair of positive samples $v_i$ and $v_j$ increases $z_i^Tz_j$; while minimizing the similarity of a pair of negative samples, say $v_i$ and $v_x$ reduces $z_i^Tz_x$.

According to Def.~\ref{def:agd-aug}, we expect an ideal encoder to generate the node embeddings such that the connected nodes in the original graph would be closer within the latent space than those unconnected nodes.

Here, we employ the Euclidean distance as the measurement in the latent space. The distance of two embeddings $z_i$ and $z_j$ is calculated as
\begin{align}
\|z_i - z_j\|&=\sqrt{\|z_i\|^2+\|z_j\|^2-2z_i^Tz_j},\nonumber \\ 
z_i^Tz_j&=\frac{\|z_i\|^2+\|z_j\|^2-\|z_i-z_j\|^2}{2}.\nonumber 
\end{align}
Moreover, we denote $P(i\sim j)$ the probability of the link existence between nodes $v_i$ and $v_j$, which is calculated by
\begin{align}
P(i\sim j)=\text{sigmoid}(z_i^Tz_j), \nonumber
\end{align}
where \text{sigmoid} is the activation function.

Suppose there is a threshold $\eta$ to indicate whether the link exists, i.e., if $\text{sigmoid}(z_i^Tz_j)> \eta$, an edge exists between nodes $v_i$ and $v_j$, and vice versa. We also assume that the embeddings are normalized, i.e., $\|z_i\|=1,\forall i$. For the connected nodes $v_i$ and $v_j$ (i.e., they form a pair of positive samples), we expect that $P(i\sim j)=\text{sigmoid}(z_i^Tz_j)>\eta$, equivalent to making $\|z_i-z_j\| < \delta$, where $\delta$ represents certain distance threshold in the latent space. Since the reconstruction and contrastive losses in Eqs.~\ref{eq_rec} and \ref{eq_contrasive} tend to maximize $z_i^Tz_j$ to make $z_i^Tz_j$ exceed link existence threshold, $\|z_i-z_j\| $ will be minimized simultaneously.
On the other hand, for the two nodes $v_i$ and $v_x$, which do not share an edge (i.e., they form a pair of negative samples), the reconstruction and contrastive losses would minimize $z_i^Tz_x$, which indicates that the distance $\|z_i-z_x\|$ in the latent space is maximized for the unconnected nodes $v_i$ and $v_x$. Please note that, according to previous work~\cite{sarkar2011theoretical}, the distance threshold $\delta$ that bounds the pairwise distances between connected nodes is finite, i.e., $\delta$ is not infinity.
}
\opt{full}{
With the reconstruction loss in Eq. \ref{eq_rec}, the model maximizes $\mathbb{E}_{q(Z|X,A)}[\log p(A|Z)]$ while minimizing $\operatorname{KL}(q(Z|X,A) \| p(Z))$. Since we predict $A_{pred}$ by calculating the inner product of $Z$, $\sigma(z_i^Tz_j)$, therefore, the model maximizes $z_i^Tz_j$ when $A_{ij}=1$ and minimizes $z_i^Tz_j$ when $A_{ij}=0$.

As for the contrastive loss in Eq. \ref{eq_contrasive}, $\text{sim}(z, z')$ represents the similarity measure between the embeddings. 
Maximizing the similarity of a pair of positive samples $v_i$ and $v_j$ increases $z_i^Tz_j$; while minimizing the similarity of a pair of negative samples, say $v_i$ and $v_x$ reduces $z_i^Tz_x$.

According to Def.~\ref{def:agd-aug}, we expect an ideal encoder to generate the node embeddings such that the connected nodes in the original graph would be closer within the latent space than those unconnected nodes.

Here, we employ the Euclidean distance as the measurement in the latent space. The distance of two embeddings $z_i$ and $z_j$ is calculated as
\begin{align}
\|z_i - z_j\|&=\sqrt{\|z_i\|^2+\|z_j\|^2-2z_i^Tz_j},\nonumber \\ 
z_i^Tz_j&=\frac{\|z_i\|^2+\|z_j\|^2-\|z_i-z_j\|^2}{2}.\nonumber 
\end{align}
Moreover, we denote $P(i\sim j)$ the probability of the link existence between nodes $v_i$ and $v_j$, which is calculated by
\begin{align}
P(i\sim j)=\text{sigmoid}(z_i^Tz_j), \nonumber
\end{align}
where \text{sigmoid} is the activation function.

Suppose there is a threshold $\eta$ to indicate whether the link exists, i.e., if $\text{sigmoid}(z_i^Tz_j)> \eta$, an edge exists between nodes $v_i$ and $v_j$, and vice versa. We also assume that the embeddings are normalized, i.e., $\|z_i\|=1,\forall i$. For the connected nodes $v_i$ and $v_j$ (i.e., they form a pair of positive samples), we expect that $P(i\sim j)=\text{sigmoid}(z_i^Tz_j)>\eta$, equivalent to making $\|z_i-z_j\| < \delta$, where $\delta$ represents certain distance threshold in the latent space. Since the reconstruction and contrastive losses in Eqs.~\ref{eq_rec} and \ref{eq_contrasive} tend to maximize $z_i^Tz_j$ to make $z_i^Tz_j$ exceed link existence threshold, $\|z_i-z_j\| $ will be minimized simultaneously.
On the other hand, for the two nodes $v_i$ and $v_x$, which do not share an edge (i.e., they form a pair of negative samples), the reconstruction and contrastive losses would minimize $z_i^Tz_x$, which indicates that the distance $\|z_i-z_x\|$ in the latent space is maximized for the unconnected nodes $v_i$ and $v_x$. Please note that, according to previous work~\cite{sarkar2011theoretical}, the distance threshold $\delta$ that bounds the pairwise distances between connected nodes is finite, i.e., $\delta$ is not infinity.
}
\end{proof}

Lemma~\ref{lemma:cluster} indicates that the distance between embeddings with connection relationship tends to be shorter after the training, i.e. $\delta$ is minimized. Intuitively, we can say that the connected nodes are more likely to be clustered together since they would be close in the latent space. Additionally, if there are $P$ clusters formed in the latent space, the distances between the embeddings within each cluster $C_p$ will also be bounded by $\delta$, where $\sup_{v_i,v_j \in C_p} \|v_i, v_j\| \leq \delta$. Based on the above analysis, the generated node embeddings still gather into latent clusters when reconstruction and contrastive losses are employed to train the link prediction model.



\opt{full}{
\begin{figure}[t]
    \centering
    \includegraphics[width=0.48\columnwidth]{distance_distr_cora.pdf}
    \includegraphics[width=0.48\columnwidth]{distance_distr_amazon_photo.pdf}
    \caption{The node-pair distance distribution in \textit{Cora} and \textit{Amazon Photo} datasets}
    \label{fig:distance_distr}
\end{figure}
}


\opt{full}{
\phead{Experimental results to illustrate Lemma~\ref{lemma:cluster}.}
To illustrate the results of Lemma~\ref{lemma:cluster}, i.e., the embeddings of the nodes which share an edge tend to cluster together under the influence of reconstruction and contrastive losses, we present additional experiments on two link prediction datasets, i.e., \emph{Cora}~\cite{wang2020microsoft} and \emph{Amazon Photo}~\cite{shchur2019pitfallsgraphneuralnetwork} in Fig.~\ref{fig:distance_distr}. Specifically, we train an autoencoder with a contrastive loss, which treats the one-hop neighbors as the positive samples by pushing them together and views the rest of the nodes as negative samples by pulling them apart, on \emph{Cora} and \emph{Amazon Photo}. We then compute the distances of the node embeddings of the i) \emph{connected nodes}, i.e., the node pairs with an edge in-between, and ii) \emph{unconnected nodes}, i.e., the node pairs without inducing an edge. 
}

\opt{full}{
\phead{Key take-away of the results in Fig.~\ref{fig:distance_distr}.}
The results are presented in Fig.~\ref{fig:distance_distr}, which indicate that the distances of the connected nodes are significantly shorter compared to those of the unconnected nodes. This validates the PLCon property discussed in Sec.~\ref{subsec:our_theoretical_part}. 

In addition, to better illustrate the clustering tendency and to understand the relationships between the node embeddings in the latent space and the corresponding connectivity in the original graph, we cluster the node embeddings in the latent space. Here, we employ TSNE to reduce the embedding to 2 dimensions for visualization, and we cluster the nodes (represented as embedding vectors in the latent space) with the $k$-means algorithm. For the connectivity of the nodes in the original graph $G$, after the nodes are clustered in the latent space, we take the nodes in each cluster and calculate their \emph{in-cluster edge density}\footnote{The in-cluster edge density calculates the edge density of the subgraph induced by nodes in the cluster.} in the original graph $G$. The in-cluster edge densities of the five clusters are all much higher ($3.5\times$ to $8.9\times$) than the edge density of the original graph $d_G$, where $d_G=0.0012$ (as shown in Fig. \ref{fig:modularity}). 
Moreover, for each cluster in the latent space, we form a community with the corresponding nodes in the original graph. We measure how well the constructed communities are formed with \textit{modularity} (a higher modularity score indicates a more pronounced community). Here, the modularity score of these communities constructed with the clustering results is $0.8317$. Typically, modularity larger than $0.3$ suggests the presence of community structures~\cite{newman2004finding}.
The significant difference between the in-cluster edge densities and $d_G$ and the high modularity validate the clustering tendency. 

}

\opt{full}{
\begin{figure}[t]
    \centering
    \includegraphics[width=0.8\columnwidth]{cluster_density.pdf}
    \caption{In-cluster edge densities of clusters in latent space (the original graph's edge density $d_G$: 0.0012, modularity: 0.8317) }
    \label{fig:modularity}
\end{figure}
}

\opt{full}{
\phead{Key take-away of the results in Fig.~\ref{fig:modularity}.}
As shown in Fig.~\ref{fig:modularity}, the embeddings of the nodes form clear clusters, while the modularity score of the communities formed with the nodes in each cluster is high. This shows that if a set of nodes incur small distances of their node embeddings in the latent space, they also present better connectivity in the original graph, and vice versa. The results support our statement outlined in Lemma~\ref{lemma:cluster} and again validate the PLCon property discussed previously.
}


\opt{full}{
\phead{Issues about the PLCon property in link prediction.}
Based on Lemma~\ref{lemma:cluster}, we conclude that connected nodes tend to gather into clusters, where each cluster represents a latent connected pattern. However, one issue might arise: suppose there are three nodes $v_1$, $v_2$, and $v_3$. If $v_1$ and $v_2$ share an edge, $v_2$ and $v_3$ share an edge, but no edge exists between $v_1$ and $v_3$ (i.e. these three nodes do not form a triangle), will the statement above be invalid?


Consider each node spans a unit sphere and other nodes located in this sphere form links. Multiple unit spheres will gather into clusters, meaning that these nodes are similar to some degree. In this case, the unit sphere can be regarded as a sub-cluster, and nodes within the same unit sphere naturally form connections due to the shared attributes. Furthermore, these similar unit spheres tend to aggregate, indicating a shared connection pattern that reflects the strength of the relationships among the nodes. Although not all the nodes within these spheres directly connect to each other due to the varying relationship strengths, and thus they might not form a clique. They typically exhibit similar connection patterns.
Furthermore, clusters could overlap, where nodes located in the overlapping area may have multiple connection patterns.
For instance, in social networks, if $A$ and $B$ are friends, while $B$ and $C$ are also friends, then $A$ and $C$ might not be directly connected. However, their mutual friendship with $B$ brings them closer than they are to other individuals with whom they share no mutual connections. This proximity due to common acquaintances differentiates their relationship from those between completely disconnected individuals, indicating a layered complexity in social network structures.

}

\phead{Dissimilarity between clusters.}
\sloppy
Based on Lemma~\ref{lemma:cluster}, we observe that connected nodes tend to cluster and can be tightened through training, enhancing the link prediction by increasing the similarity and link probability of closely located node pairs in the latent space. The remaining task is to understand how these clusters assist in distinguishing node pairs that are not connected. To this end, we introduce Lemma~\ref{lemma:inter_similarity}, which states how nodes from different clusters exhibit dissimilarities. 

Recall that $\gamma$, $\alpha$, and $\hat{d}$ are the parameters for the $(\alpha,\gamma,\hat{d})$-augmentation defined in Def.~\ref{def:agd-aug}, while $B$ denotes the feature dimension, $\epsilon$ is the parameter in $\epsilon$-close node set, and $R_\epsilon$ is the bounding term in Theorem~\ref{thm:downstream_error}. In the following, to simplify the equations, we let $\rho_{\text{max}}(\alpha, \gamma, \hat{d}, \epsilon) = 2(1 - \alpha) + \max_{\forall \ell} \left( \frac{2R_\epsilon}{p_\ell} + \frac{M\alpha \gamma \sqrt{B}}{r\sqrt{\hat{d}_{\min}^\ell}} \right) + \frac{2\alpha \epsilon}{r}$, where $r$ is the normalization term of the embeddings. Also, $M$ is the constant of $M$-Lipschitz continuity which limits the variation of the function $f$ to at most $M$ times the changes in the input; $p_p = \mathbb{P}[v_i \in C_p]$ is the probability of node $v_i$ belonging to cluster $C_p$; 
$\Delta_\mu = 1 - \min_{p \in [P]} \frac{\|\mu_p\|^2}{r^2}$, where $P$ is the number of clusters in the latent space and $\mu_p$ is the center of the cluster $C_p$. 

\begin{lemma} 
\textbf{Inter-cluster similarity.}
\label{lemma:inter_similarity}
\sloppy
For two node embeddings from different latent clusters, if the embeddings can be correctly encoded by the model, then their similarity (inner product) is upper-bounded by $r^2 \left(1 - \rho_{\text{max}}(\alpha, \gamma, \hat{d}, \epsilon)- \sqrt{2 \rho_{\text{max}}(\alpha, \gamma, \hat{d}, \epsilon)} - \frac{\Delta \mu}{2} \right)+{E}_{ij}$, where $\text{E}_{ij}= 8r \left(1 - \alpha \left(1 - \frac{\epsilon}{2r} - \frac{M }{4r}\gamma \left( \frac{B}{\hat{d}^p_{\min}} \right)^{\frac{1}{2}}\right) + \frac{R_\epsilon}{p_p}\right) + 16r^2 \left(1 - \alpha \left(1 - \frac{\epsilon}{2r} - \frac{M}{4r}\gamma \left( \frac{B}{\hat{d}^p_{\min}} \right)^{\frac{1}{2}}\right) + \frac{R_\epsilon}{p_p}\right)^2$,  $d_{\text{min}}^p$ denotes the minimum node degree of the nodes in the cluster $C_p$, and $r$ is the normalization term of the embeddings.
\end{lemma}

\begin{proof}
\opt{short}{Detailed proof is in the online full version~\cite{rep_material}. }
\opt{full}{
Let $\mu_\ell$ and $\mu_p$ denote the cluster centers of the clusters $C_\ell$ and $C_p$, respectively. Let $z_i$ and $z_j$ denote the two different node embeddings from the clusters $C_\ell$ and $C_p$, respectively. Based on the earlier discussion regarding cluster-based dissimilarity and Lemma~\ref{lemma:cluster}, we can model each node embedding as a combination of its latent cluster center and a deviation vector. Specifically, we define the embedding $z_i$ of node $v_i \in C_\ell$ as 
\begin{equation*}
\begin{aligned}
z_i = \mu_\ell + e_i; z_j = \mu_p + e_j,
\end{aligned}
\end{equation*}
where $e_i$ and $e_j$ are the bias vectors from the embeddings to the corresponding cluster centers, respectively.
Now, we employ inner product to calculate the similarity of $z_i$ and $z_j$, i.e., $z_i^T z_j = (\mu_\ell + e_i)^T (\mu_p + e_j) = \mu_\ell^T \mu_p + \mu_\ell^T e_j + e_i^T \mu_p + e_i^T e_j$.
Please note that the upper-bound on the inner product of the two centers of latent clusters $\mu_\ell$ and $\mu_p$ is
$\mu_{\ell}^T \mu_p < r^2 \left(1 - \rho_{\max}(\alpha, \gamma, \hat{d}, \epsilon) - \sqrt{2 \rho_{\max}(\alpha, \gamma, \hat{d}, \epsilon)} - \frac{\Delta \mu}{2} \right)$~\cite{Wang2022}.

Here, $\|e_i\|$ and $\|e_j\|$ can be approximated and bounded by the following inequality~\cite{huang2023towards}, where $\mathcal{T}(v)$ is the augmentation set of the potential positive samples generated from node $v$.
\begin{align}
\|e_i\| &= \mathbb{E}_{v \in C_p} \mathbb{E}_{z_i=f(v_i),v_i \in \mathcal{T}(v)} \| z_i - \mu_p \| \nonumber \\
& \leq 4r \left( 1 - \alpha \left( 1 - \frac{\epsilon}{2r} - \frac{M\gamma}{4r}\left( \frac{B}{\hat{d}^p_{\min}} \right)^{\frac{1}{2}} \right) + \frac{R_{\epsilon}}{p_p}\right) \label{eq_ei}
\end{align}

Using Cauchy-Schwartz inequality and applying Eq. \ref{eq_ei}, we approximate $ e_i^T e_j$ as follows.
\begin{align}
e_i^T e_j &\leq \|e_i\| \|e_j\|,\\ \nonumber
e_i^T e_j &\leq 16r^2 \left(1 - \alpha \left(1 - \frac{\epsilon}{2r} - \frac{M \gamma }{4r}\left( \frac{B}{\hat{d}^p_{\min}} \right)^{\frac{1}{2}}\right) + \frac{R_\epsilon}{p_p}\right)^2. \nonumber
\end{align}

Therefore, the final upper bound $v_i^T v_j $ is derived as follows.
\begin{align}
\label{upperbound_vivj}
\mathbf{z}_i^T \mathbf{z}_j & \leq r^2 \left(1 - \rho_{\text{max}}(\alpha, \gamma, \hat{d}, \epsilon) - \sqrt{2 \rho_{\text{max}}(\alpha, \gamma, \hat{d}, \epsilon)} - \frac{\Delta \mu}{2} \right) \nonumber \\
&+ 2{\cdot}4r \left( 1 - \alpha \left( 1 - \frac{\epsilon}{2r} - \frac{M_\gamma }{4r}\left( \frac{B}{\hat{d}_{\min}^p} \right) \right)^{\frac{1}{2}} + \frac{R_\epsilon}{p_p} \right) \nonumber \\
&+ 16r^2 \left( 1 - \alpha \left( 1 - \frac{\epsilon}{2r} - \frac{M_\gamma}{4r} \left( \frac{B}{\hat{d}_{\min}^p} \right) \right) + \frac{R_\epsilon}{p_p} \right). 
\end{align}
}
\opt{online}{
Let $\mu_\ell$ and $\mu_p$ denote the cluster centers of the clusters $C_\ell$ and $C_p$, respectively. Let $z_i$ and $z_j$ denote the two different node embeddings from the clusters $C_\ell$ and $C_p$, respectively. Based on the earlier discussion regarding cluster-based dissimilarity and Lemma~\ref{lemma:cluster}, we can model each node embedding as a combination of its latent cluster center and a deviation vector. Specifically, we define the embedding $z_i$ of node $v_i \in C_\ell$ as 
\begin{equation*}
\begin{aligned}
z_i = \mu_\ell + e_i; z_j = \mu_p + e_j,
\end{aligned}
\end{equation*}
where $e_i$ and $e_j$ are the bias vectors from the embeddings to the corresponding cluster centers, respectively.
Now, we employ inner product to calculate the similarity of $z_i$ and $z_j$, i.e., $z_i^T z_j = (\mu_\ell + e_i)^T (\mu_p + e_j) = \mu_\ell^T \mu_p + \mu_\ell^T e_j + e_i^T \mu_p + e_i^T e_j$.
Please note that the upper-bound on the inner product of the two centers of latent clusters $\mu_\ell$ and $\mu_p$ is
$\mu_{\ell}^T \mu_p < r^2 \left(1 - \rho_{\max}(\alpha, \gamma, \hat{d}, \epsilon) - \sqrt{2 \rho_{\max}(\alpha, \gamma, \hat{d}, \epsilon)} - \frac{\Delta \mu}{2} \right)$~\cite{Wang2022}.

Here, $\|e_i\|$ and $\|e_j\|$ can be approximated and bounded by the following inequality~\cite{huang2023towards}, where $\mathcal{T}(v)$ is the augmentation set of the potential positive samples generated from node $v$.
\begin{align}
\|e_i\| &= \mathbb{E}_{v \in C_p} \mathbb{E}_{z_i=f(v_i),v_i \in \mathcal{T}(v)} \| z_i - \mu_p \| \nonumber \\
& \leq 4r \left( 1 - \alpha \left( 1 - \frac{\epsilon}{2r} - \frac{M\gamma}{4r}\left( \frac{B}{\hat{d}^p_{\min}} \right)^{\frac{1}{2}} \right) + \frac{R_{\epsilon}}{p_p}\right) \label{eq_ei}
\end{align}

Using Cauchy-Schwartz inequality and applying Eq. \ref{eq_ei}, we approximate $ e_i^T e_j$ as follows.
\begin{align}
e_i^T e_j &\leq \|e_i\| \|e_j\|,\\ \nonumber
e_i^T e_j &\leq 16r^2 \left(1 - \alpha \left(1 - \frac{\epsilon}{2r} - \frac{M \gamma }{4r}\left( \frac{B}{\hat{d}^p_{\min}} \right)^{\frac{1}{2}}\right) + \frac{R_\epsilon}{p_p}\right)^2. \nonumber
\end{align}

Therefore, the final upper bound $v_i^T v_j $ is derived as follows.
\begin{align}
\label{upperbound_vivj}
\mathbf{z}_i^T \mathbf{z}_j & \leq r^2 \left(1 - \rho_{\text{max}}(\alpha, \gamma, \hat{d}, \epsilon) - \sqrt{2 \rho_{\text{max}}(\alpha, \gamma, \hat{d}, \epsilon)} - \frac{\Delta \mu}{2} \right) \nonumber \\
&+ 2{\cdot}4r \left( 1 - \alpha \left( 1 - \frac{\epsilon}{2r} - \frac{M_\gamma }{4r}\left( \frac{B}{\hat{d}_{\min}^p} \right) \right)^{\frac{1}{2}} + \frac{R_\epsilon}{p_p} \right) \nonumber \\
&+ 16r^2 \left( 1 - \alpha \left( 1 - \frac{\epsilon}{2r} - \frac{M_\gamma}{4r} \left( \frac{B}{\hat{d}_{\min}^p} \right) \right) + \frac{R_\epsilon}{p_p} \right). 
\end{align}
}
\end{proof}

Lemma~\ref{lemma:inter_similarity} gives the upper bound of the similarity between the nodes from different clusters. It implies that once we minimize 
\opt{short}{$r^2 \left(1 - \rho_{\text{max}}(\alpha, \gamma, \hat{d}, \epsilon)- \sqrt{2 \rho_{\text{max}}(\alpha, \gamma, \hat{d}, \epsilon)} - \frac{\Delta \mu}{2} \right)+{E}_{ij}$}
\opt{full}{the Right-Hand-Side of Eq. \ref{upperbound_vivj}}
\opt{online}{the Right-Hand-Side of Eq. \ref{upperbound_vivj}}, the probability of the link existence between the nodes in different clusters will also be minimized. Since the encoder puts nodes that have potential connections together, minimizing the probability of forming links between dissimilar node pairs is equivalent to reducing the probability of incorrectly forming a link between node pairs that should not form a link.

\phead{Summary.}
From the analysis above, we conclude that embeddings trained by the contrastive link prediction model exhibit the PLCon property, indicating that embeddings with potential connections tend to cluster in the latent space. Furthermore, there is an upper bound on the clustering error rate of the encoder. We also establish an upper bound on the dissimilarity of embeddings from different clusters after the encoder training, and minimizing this bound reduces the likelihood of the encoder clustering unconnected nodes together. The primary strategies for enhancing the link prediction performance (i.e., reducing the clustering error rate) involve either i) increasing the proportion of the main part in each cluster, thereby resulting in a larger $\alpha$, or ii) making the embedding more concentrated. While the former is addressed through the alignment loss, as most contrastive learning approaches focus on, our proposed approach further focuses on making the learned embeddings more concentrated, which has not been well explored in the link prediction task. 

It is worth noting that while previous works have analyzed the concentration and error bounds of contrastive learning in node classification tasks, these results cannot be directly transferred to link prediction due to fundamental differences in task formulation. Our work is \emph{the first to bridge this gap by theoretically analyzing the relationship between node degree and link prediction performance}. This theoretical contribution fills an important void in the literature and aligns with recent studies~\cite{xiao2023simple,zhou2023combating} that emphasize the value of theoretical analysis in guiding practical algorithm design.

In the following, we formalize our theoretical findings into Corollary~\ref{cor:direction}, which guides the design of data augmentation strategies in our proposed approach in Sec.~\ref{sec:algo}. 

\begin{corollary}
\label{cor:direction}
To improve the performance of the link prediction model employing the contrastive loss, enhancing the concentration of embeddings is critical. This involves minimizing the upper bound $\gamma \left(\frac{B}{\hat{d}^{p}_{\min}}\right)^{\frac{1}{2}}$ in $\sup_{v_i,v_j \in C^0_p} d_\mathcal{T}(v_i, v_j) \leq \gamma \left(\frac{B}{\hat{d}^{p}_{\min}}\right)^{\frac{1}{2}}$, where $\gamma \in (0,1]$. Essentially, this requires decreasing $B$ or increasing $\hat{d}^{p}_{\min}$, where $B$ is the dimension of features, and $\hat{d}^{p}_{\min}$ is the minimum node degree in the main part $C_p^0$ of each cluster.
\end{corollary}





\section{Algorithm Design}
\label{sec:algo}

\begin{figure*}[t]
    \centering
    \subfloat[]
    {%
        \includegraphics[width=1.35\columnwidth]{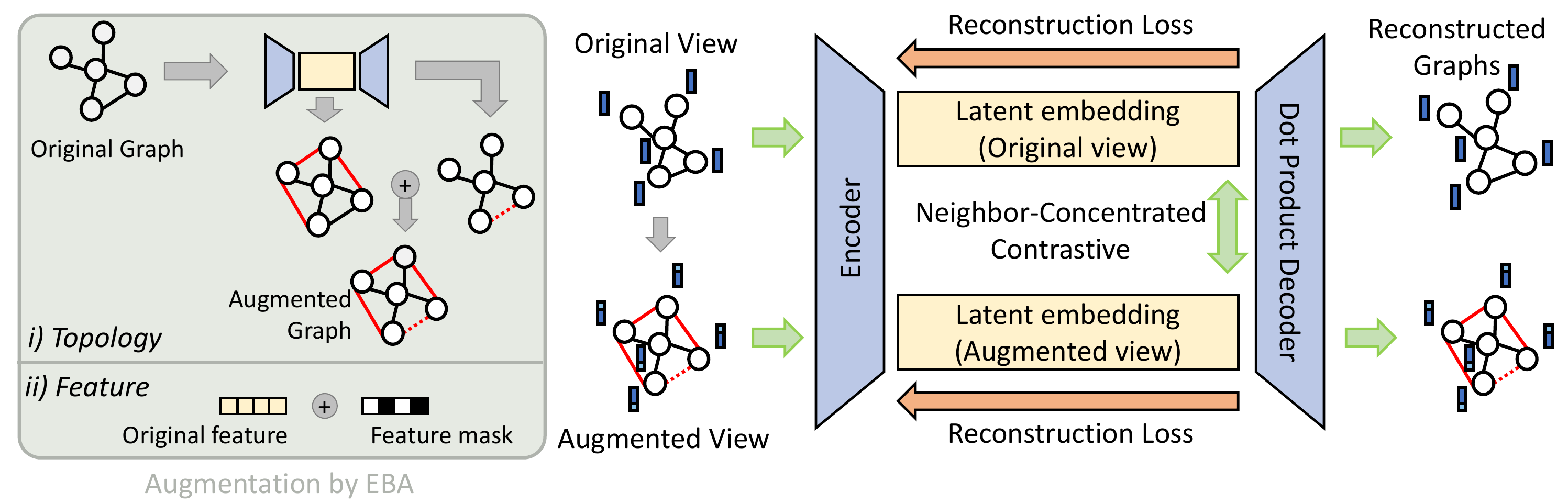}
        \label{fig:eba}
    }
    \hfill
    \subfloat[]
    {%
        \includegraphics[width=0.73\columnwidth]{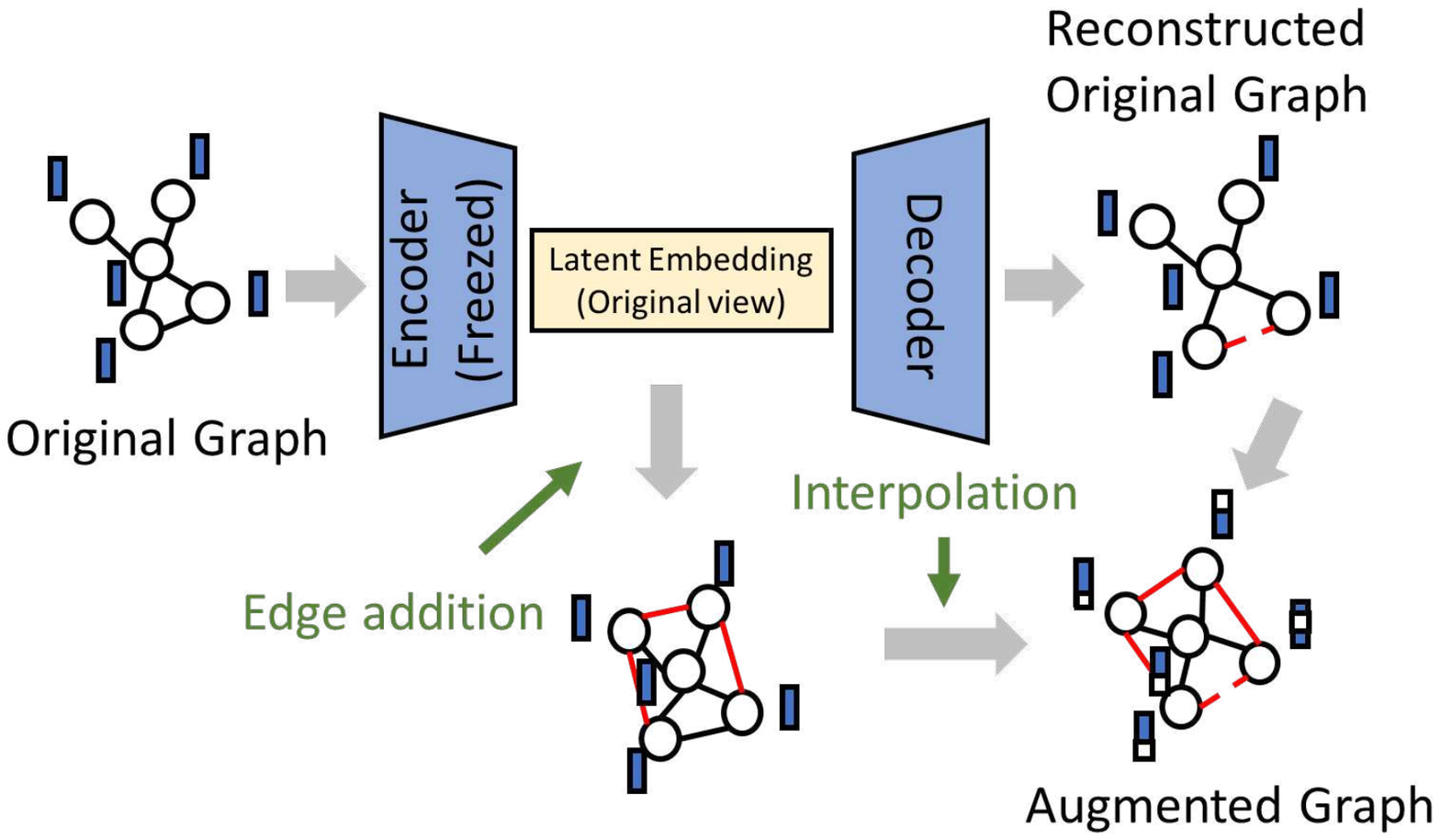}
        \label{fig:eba_process}
    }
    \subfloat[]
    {%
        \includegraphics[width=0.73\columnwidth]{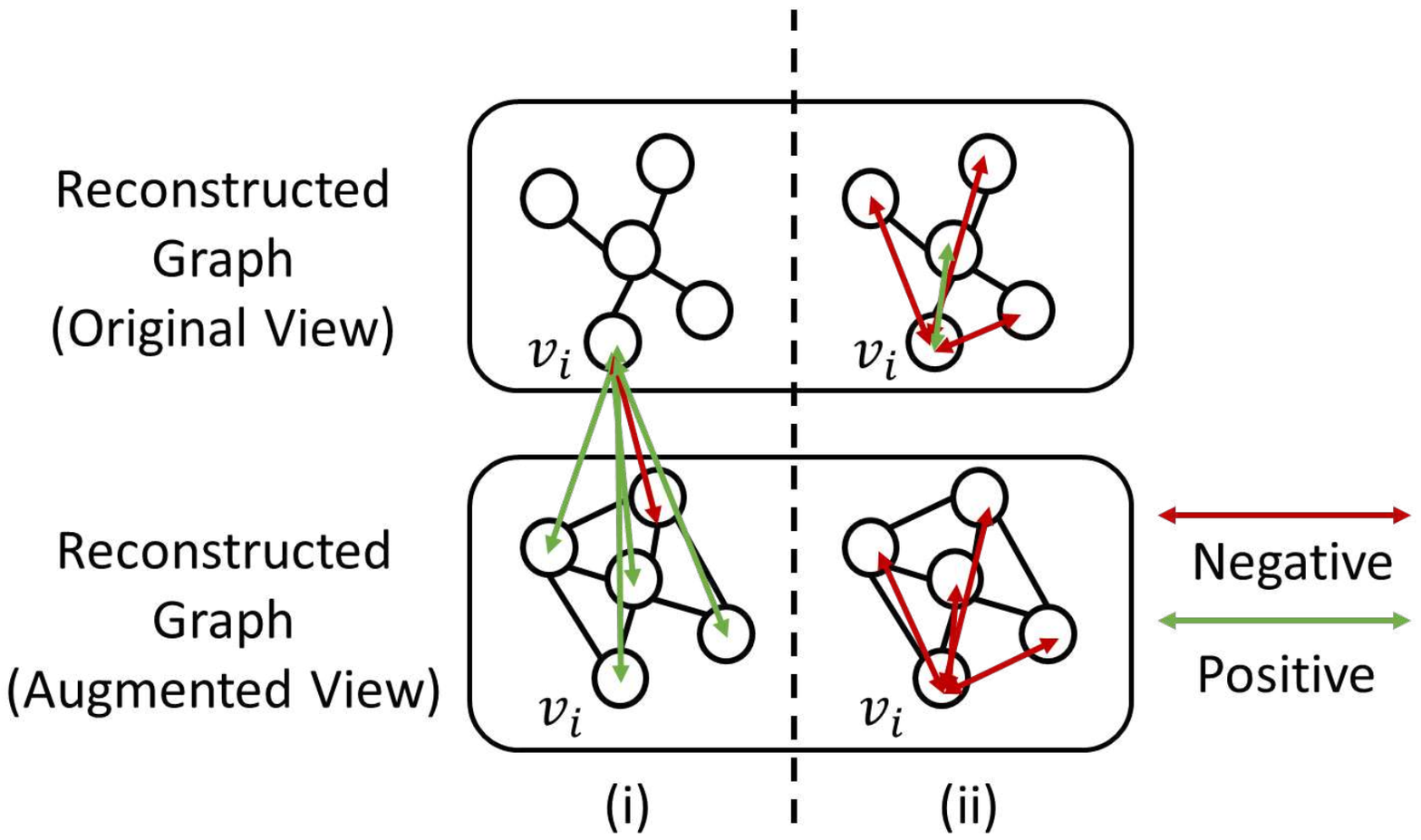}
        \label{fig:contrastive}
    }
    \caption{Illustration of the proposed \texttt{CoEBA}: (a) Overview of the full pipeline; (b) Illustration of \emph{Edge Balancing Augmentation (EBA)}; (c) The proposed neighbor-concentrated contrastive objective }
    \label{fig:eba_process_contrastive}
\end{figure*}

Fig.~\ref{fig:eba} illustrates the overall structure of our proposed approach, \emph{Contrastive Link Prediction with Edge Balancing Augmentation} (\texttt{CoEBA}), which consists of an encoder, a decoder, and a graph augmentation module featuring our proposed \emph{Edge Balancing Augmentation (EBA)}.
The input of our model is a graph $G=(V, E, X)$, where the nodes and edges, $(V, E)$, are represented as an adjacency matrix $A$. For an element $a_{ij}\in A$, $a_{ij}=1$ if and only if there exists an edge $(v_i,v_j)$. The graph augmentation is performed on the original graph $G=(V,E,X)$, which transforms it into the augmented graph $G^{\prime}=(V^{\prime},E^{\prime},X^{\prime})$. These graphs, i.e., $G$ and $G^{\prime}$, are then fed into the encoder and decoder, resulting in a reconstructed edge set $E_{pred}$. This reconstructed edge set, which is represented as an adjacency matrix $A_{pred}$, is the model outcome.

More specifically, at the beginning, we generate an augmented graph with the proposed \emph{Edge Balancing Augmentation} (EBA), which incorporates both topology and feature augmentations. Then, both the original and augmented graphs are processed through a graph autoencoder to obtain their embeddings. We employ the proposed \emph{neighbor-concentrated contrastive loss} to contrast the learned embeddings of the original and augmented graphs. Finally, the decoder reconstructs these embeddings into a reconstructed graph, which is then used to optimize the model via reconstruction loss. 


\subsection{Edge Balancing Augmentation (EBA)}

Based on the analysis in Sec.~\ref{subsec:our_theoretical_part}, especially Corollary~\ref{cor:direction}, the key to improving the generalization ability of contrastive learning for link prediction lies in minimizing the theoretical upper bound $\gamma \left( \frac{B}{\hat{d}^{p}_{\min}} \right)^{1/2}$. According to our analysis in Sec.~\ref{subsec:our_theoretical_part}, increasing the minimum node degree makes nodes with the same connection pattern closer to each other. Motivated by this, we propose \emph{Edge Balancing Augmentation (EBA)} strategy to explicitly adjust the degree distribution. 
This process effectively increases the minimum node degree within a latent cluster $\hat{d}^{p}_{\min}$, thereby reducing clustering errors. Fig.~\ref{fig:eba_process} illustrates the EBA procedure, where the node degrees are balanced and the minimum node degree is increased after the augmentation.


\phead{High-level concept of EBA.} Recall that our goal is to maximize the minimum node degree within the clusters. However, the complex structure of the graph inherently limits the extent to which we can increase edges without compromising the semantic structure of the original graph. Excessive edge augmentation might distort the original graph's structure and deteriorate the model performance. Therefore, when designing augmentation strategies, two key issues should be addressed simultaneously: i) preserving the structure of the original graph and ii) increasing the minimum node degree.

Aiming the first issue above and to preserve the structure of the original graph, we augment only a small portion of the original graph to construct the augmented graph, where the portion of edge augmentation is controlled by the \emph{neighbor removal ratio $r_m\in [0,1]$} and \emph{neighbor addition ratio $r_a\in [1,0]$}. As their names suggest, our augmentation includes both neighbor removal and addition, which could be viewed as a \emph{graph interpolation}. 
More specifically, for node $v$ in the original graph, we augment $v$'s neighbors (denoted as $N_G(v)$) in the augmented graph with the following two steps: a) \emph{Noise removal}, where $r_m{\cdot}|N_G(v)|$ of $v$'s incident edges are removed if they are not confidently predicted by the link prediction model, i.e., these edges can be viewed as the noise; b) \emph{Edge addition among similar nodes}, where $r_a{\cdot}|N_G(v)|$ ($r_a>r_m$) new edges are added when edges' two endpoints are close in the latent space. These added edges reinforce the graph structure and enhance the message-passing of the GNN.  

To address the second issue, for each node $v$, let $x=r_a\cdot |N_G(v)|$ denote the number of new neighbors to be added to node $v$ for the strategy: \emph{edge addition among similar nodes}. We extract top-$x$ nodes most similar to $v$ based on their embeddings and add these nodes as the new neighbors of $v$ in the augmented graph. Here, the number $x$ represents a trade-off between increasing the node degree of $v$ and maintaining the original graph structure. For example, adding too few new neighbors (i.e., setting a too small $x$) may not significantly decrease $\gamma \left(\frac{B}{\hat{d}^{p}_{\min}}\right)^{\frac{1}{2}}$, leading to large error rate; However, adding too many new neighbors (i.e., setting a too large $x$) leads to the significant increase in the alignment loss $\mathbb{E}_{v_i} \mathbb{E}_{\hat{G_i^1}, \hat{G_i^2} \in \mathcal{T}(G_i)} \left\| f(\hat{G_i^1}) - f(\hat{G_i^2}) \right\|$, which again degrades the model performance.


In the following, we describe how EBA works. Recall that $X$ and $A$ are the feature matrix and adjacency matrix of the input graph $G$, respectively, and let $f$ denote the encoder. We first perform a \emph{warm-up training} to train the encoder $f$ and decoder $\widehat{f}$ for a few epochs (in this paper, we warm up the model with 200 epochs) to obtain the initial model and the corresponding node embeddings. This ensures that the node embeddings are adequately refined and stable, making them suitable to perform EBA.

We first obtain the embedding $Z$ by inputting $X$ and $A$ to the encoder $f$, i.e., $Z = f(X,A)$, where the model parameter set of $f$ is frozen during EBA. Then, we perform \emph{noise removal} by pruning the edges which are not confidently predicted by the model. 
Specifically, we employ the initial model to obtain the reconstructed adjacency matrix $A^{init}_{pred}$, of which each element $a_{ij}\in [0,1]$  represents the probability of the existence of edge $(v_i,v_j)$ predicted by the model.

The augmented graph $G^{\prime}=(V^{\prime},E^{\prime},X^{\prime})$ is initialized as a copy of the original graph $G=(V,E,X)$. Then, for each node $v\in V^{\prime}$, we rank each neighbor $u\in N_G(v)$ of $v$ according to the initial model prediction, i.e., $a_{uv}\in A^{init}_{pred}$. Based on the neighbor removal ratio $r_m$, we prune the $r_m\cdot |N_G(v)|$ neighbors with the smallest predicted values. We preserve the remaining neighbors as part of the new neighbor in the augmented graph. Then, we perform \emph{edge addition among similar nodes}, as described above, to link $v$ to the top-$(r_a\cdot |N_G(v)|)$ most similar nodes based on their embeddings in the augmented graph. Now, we have the new $V^{\prime}$ and $E^{\prime}$.

In addition to augmenting the edges in EBA, we also perform feature augmentation by random feature masking, which is an effective strategy widely adopted by contrastive learning on graphs~\cite{zhu2021, you2020}. 
More specifically, following previous research~\cite{zhu2021}, we construct a random mask $M$ to remove a subset of features of the original feature matrix $X$ and produce the augmented feature matrix $X^{\prime}$. So, we now have completed the generation of the augmented graph $G^{\prime}=(V^{\prime},E^{\prime},X^{\prime})$.


\subsection{Encoding and Decoding}
Our objective is to reconstruct the edge set $E_{pred}$ to approximate the ideal edge set, $E_{ideal}$, as closely as possible. 
To ensure compatibility with standard autoencoder-based link prediction frameworks, we adopt analogous model strategies. Following the assumption in VGAE~\cite{kipf2016}, we treat the embedding $Z$ as a latent generator of the input adjacency matrix 
$A$, responsible for the generation of all edges.
Following previous works~\cite{yang2023, tan2022, ahn2021}, the inference model $q(Z|X, A)$ includes a two-layer GNN and employs an MLP with batch normalization as the projection mechanism as follows:
\begin{equation}\label{gae1}
q(Z \mid X, A) = \prod_{i=1}^{N} \mathcal{N}(z_i \mid \mu_i, \text{diag}(\sigma_i^2)),
\end{equation}
where $N$ is the number of nodes, while $\hat{\mu}_i$ and $\sigma_i$ denote the mean and standard deviation vectors of the encoder's output for node $v_i$, respectively. Here, $\mathcal{N}$ is the multivariate normal distribution with diagonal covariance, while $z_i$ and $z_j \in Z$ are the embeddings of nodes $v_i$ and $v_j$, respectively.

The generative model $p(A|Z)$ is referred to as our probability decoder to reconstruct the graph structure.
\begin{equation}\label{gae3}
p(A \mid Z) = \prod_{i=1}^{N} \prod_{j=1}^{N} \text{sigmoid}(z_i \cdot z_j).
\end{equation}
The model is optimized by the reconstruction loss, which is the negative variational lower bound \cite{kipf2016}.
\begin{equation}\label{recon_loss}
\mathcal{L}_{\text{recon}} = -\mathbb{E}_{q(Z \mid X,A)} [\log p(A \mid Z)] + \text{KL}(q(Z \mid X, A) \parallel p(Z)),
\end{equation}
where $\text{KL}[ \cdot \parallel \cdot ]$ is the Kullback-Leibler divergence between the two distributions, and $p(Z)$ is a Gaussian prior.

After the augmentation, we obtain the embedding $Z$ from the original graph by feeding the original graph and features into the encoder $f$, i.e., $Z=f(X,A)$. We obtain the embedding $Z^{\prime}$ from the augmented graph by putting the augmented graph and augmented features into the same encoder $f$, i.e., $Z^{\prime}=f(X^{\prime},A^{\prime})$.

\subsection{The Proposed New Contrastive Losses}

\phead{Neighbor-concentrated contrastive losses.}
To further improve the learned node embedding, we incorporate contrastive learning with our autoencoder approach as shown in Fig.~\ref{fig:contrastive}. Inspired by the concept of neighbor contrastive~\cite{Shen2023}, we propose the \emph{neighbor-concentrated contrastive losses} to improve the concentration of the node embeddings that their corresponding nodes in the graph tend to connect.  In previous works~\cite{oord2019, zhu2020}, for any anchor node $v_i$, they usually treat other nodes in the same graph as negative pairs with $v_i$. However, this causes the embedding of $v_i$ and its neighbors to be pushed apart, which hurts the performance of link prediction. 

To tackle this issue, for an anchor node $v_i$ in the original graph $G$, its positive pairs include $v_i$ itself and its one-hop neighbors $N_G(v_i)$, while the other nodes are the negative pairs. Further, in the augmented graph $G^{\prime}$, for an anchor node $v_i$, we only include $v_i$ itself as the positive pair while the rest nodes form the negative pairs with $v_i$. This is because the augmented graph is generated with our augmentation strategy, which may unavoidably include biases and uncertainty. For example, at the top of Fig.~\ref{fig:contrastive}-(ii), the node $v_i$ from the reconstructed original graph treats its neighbors as positive pairs (green arrows), while taking the other unconnected nodes as negative pairs (red arrows). At the bottom of Fig.~\ref{fig:contrastive}-(ii), the node $v_i$ from the reconstructed augmented graph takes all the other nodes as negative pairs (red arrows).

The first set of neighbor-concentrated contrastive losses, i.e., \emph{within-CL losses}, includes the losses on the augmented graph ($\mathcal{L}_{\text{cnst}}^{\text{aug}}(v_i)$) and original graph ($\mathcal{L}_{\text{cnst}}^{\text{ori}}(v_i)$), as  specified below.
\begin{align}
\footnotesize
\mathcal{L}_{\text{cnst}}^{\text{aug}}(v_i) &= -\log \frac{e^{\text{sim}(z_i^{\prime}, z_i^{\prime}) / \tau}}{\sum_{j=1}^N e^{\text{sim}(z_i^{\prime}, z_j^{\prime}) / \tau}},\label{intra_aug} \\
\mathcal{L}_{\text{cnst}}^{\text{ori}}(v_i) &= -\log \frac{\sum_{z_j \in \{z_i\}\cup N_{G}(z_i)} e^{\text{sim}(z_i, z_j) / \tau}}{\sum_{j=1}^N e^{\text{sim}(z_i, z_j) / \tau}},\label{intra_ori}
\end{align}
where $\text{sim}(\cdot)$ is the similarity function of the embeddings, and $\tau$ is the temperature hyperparameter. $z_i$ and $z_j$ are the node embeddings of nodes $v_i$ and $v_j$ from the embedding $Z$ of the original graph, respectively. $z_i^{\prime}$ and $z_j^{\prime}$ are the node embeddings of nodes $v_i$ and $v_j$ from the embedding $Z^{\prime}$ of the augmented graph, respectively.

For the contrastive loss between the original and augmented graphs, the goal is to contrast the nodes between the original and augmented graphs. Similar to the concept of \emph{within-CL losses}, we also take the neighbor structure as positive samples. That is, for node $v_i$ in the original graph $G$, node $v_i$ and its neighbors in the augmented graph $G^{\prime}$ form the positive samples for node $v_i$, while the other nodes are the negative samples. 
Take Fig.~\ref{fig:contrastive}-(i) as an example. The positive pairs of node $v_i$ in the reconstructed original graph are the corresponding node of $v_i$ and its neighbors in the reconstructed augmented graph (green arrows). The other unconnected nodes are negative pairs (red arrows).

The set of losses in the proposed neighbor-concentrated contrastive losses captures the relationships between the graphs and is named \emph{btn-CL loss} ($\mathcal{L}_{\text{btn}}$), which is defined as follows. 
\begin{align}
\mathcal{L}_{\text{btn}}(v_i) &= -\log \frac{\sum_{z_j^{\prime} \in \{z_i^{\prime}\}\cup N_{G^{\prime}}(z_i^{\prime})} e^{\text{sim}(z_i, z_j^{\prime}) / \tau}}{\sum_{j=1}^N e^{\text{sim}(z_i, z_j^{\prime}) / \tau}},\nonumber \\
\mathcal{L}_{\text{btn}} &= \frac{1}{N} \sum_{i=1}^N \mathcal{L}_{\text{btn}}(v_i). \label{inter}
\end{align}

\phead{Overall loss.} We train our model using the overall loss that includes the above reconstruction loss (Eq. \ref{recon_loss}), along with the proposed neighbor-concentrated contrastive losses, which integrate the \emph{within-CL losses} (Eqs. \ref{intra_aug} and \ref{intra_ori}), and \emph{btn-CL loss} (Eq. \ref{inter}). The overall loss is formulated below.
\begin{equation}
\begin{split}
\mathcal{L}_{all} &= \mathcal{L}_{btn} + \lambda_1{\cdot}(\mathcal{L}_{recon}^{ori} + \mathcal{L}_{recon}^{aug} ) + \lambda_2{\cdot}\mathcal{L}_{cnst}^{ori} + \lambda_3{\cdot}\mathcal{L}_{cnst}^{aug},
\end{split}
\label{overall_loss}
\end{equation}
where $\lambda_1$, $\lambda_2$, and $\lambda_3$ are the hyperparameters that control the balance between these loss terms.



\begin{algorithm}[t]
\caption{Training pipeline of \texttt{CoEBA}}\label{algo:pipeline}
\begin{algorithmic}[1]
\small
\State \textbf{Input:} $G=(V,E,X)$
\State \textbf{Output:} Prediction result $A_{\text{pred}}$
\For{$e$ in the range of epoch number}
    \If{$e \% t = 0$}
        \State \textbf{with} no gradient \textbf{do}
        \State \hspace*{\algorithmicindent} $Z = f^{freezed}_{\theta}(X, A)$
        \State Prune the $r_m \cdot |N_G(v)|$ neighbors with the smallest predicted values, $\forall v \in G$
        \State Link $v$ to top-$(r_a\cdot |N_G(v)|)$ most similar nodes, $\forall v \in G$
        \State Obtain the augmented graph $G^{\prime}$ 
        \State Obtain the augmented feature $X^{\prime}$ by feature augmentation
    \EndIf
    \State $Z = f_{\theta}(X, A)$ \Comment{Original graph reconstruction}
    \State $A_{pred} = decoder(Z)$
    \State $Z^{\prime} = f_{\theta}(X^{\prime}, A^{\prime})$
    \State $A^{\prime}_{pred} = decoder(Z^{\prime})$ \Comment{Augmented graph reconstruction}
    \State $\mathcal{L}_{recon}^{ori} \leftarrow$ Reconstruction loss of $A_{pred}$ in Eq. \ref{recon_loss}
    \State $\mathcal{L}_{recon}^{aug} \leftarrow$ Reconstruction loss of $A_{pred}^{\prime}$ in Eq. \ref{recon_loss}
    \State $\mathcal{L}^{aug}_{cnst}, \mathcal{L}^{ori}_{cnst} \leftarrow$ \emph{Within-CL losses} in Eqs. \ref{intra_aug}, \ref{intra_ori}
    \State $\mathcal{L}_{btn} \leftarrow$ \emph{btn-CL loss} in Eq. \ref{inter}
    \State $\mathcal{L}_{all} \leftarrow$ Overall loss in Eq. \ref{overall_loss}
    \State Update $\theta$ by minimizing $\mathcal{L}_{all}$
\EndFor
\State \textbf{with} no gradient \textbf{do}\Comment{end of training}
\State \hspace*{\algorithmicindent}
$Z = f_{\theta}(X, A)$
\State \hspace*{\algorithmicindent}
${A}_{pred} = decoder(Z)$
\State \Return ${A}_{pred}$
\end{algorithmic}
\end{algorithm}

\phead{Time complexity.} Denote the dimension of input $X$ as $N \times D_f$, where $N$ is the number of nodes and $D_f$ is the dimension of a node's embedding. To construct graphs, \texttt{CoEBA} needs to calculate the distance between all pairs of nodes and choose $k$ neighbors for every node. Thus, the time complexity is $O(N^2 \times D_f + kN)$, where $O(D_f)$ is for distance calculation. To augment new graphs, \texttt{CoEBA} finds the deleted nodes in $O(N)$, samples new neighbors in $O(r_a \times N)$ (where $r_a$ is the neighbor addition ratio), and creates the adjacency matrices of augmented graphs in $O(N^2)$. Thus, the time complexity is $O(N^2 \times D_f)$. The pseudo-code of \texttt{CoEBA} is shown in Algorithm~\ref{algo:pipeline}. 
Although the complexity is $O(N^2 \times D_f)$, \texttt{CoEBA} is still efficient in running time, as detailed in Sec.~\ref{sec:exp}.

\section{Experimental Results}
\label{sec:exp}

 \opt{full}{
\phead{Datasets.}
We conduct our experiments on 9 widely-adopted benchmark datasets to evaluate the link prediction task. 
First, we test on the large-scale \emph{ogbl-collab}~\cite{wang2020microsoft} dataset with 235K nodes and 1.28M edges.
Other datasets are also collected from the real world across different domains, including 
2 co-purchase networks, \emph{Amazon Computers}~\cite{shchur2019pitfallsgraphneuralnetwork} (or \emph{Amazon Com.} for short) with 13K nodes and 491K edges, and \emph{Amazon Photo}~\cite{shchur2019pitfallsgraphneuralnetwork} (or \emph{Amazon P.} for short), 3 citation networks, \emph{Cora}~\cite{mccallum2000automating}, \emph{Cora\_ML}~\cite{mccallum2000automating}, and \emph{Citeseer}~\cite{giles1998citeseer}, and 3 webpage datasets, \emph{Cornell}~\cite{WebKB}, \emph{Texas}~\cite{WebKB}, and \emph{Wisconsin}~\cite{WebKB}.  
The statistics of the above datasets are summarized in Table~\ref{tab:dataset_statistics}. }
\opt{short}{
\phead{Datasets.}
We conduct our experiments on 8 widely adopted benchmark datasets to evaluate the link prediction task. 
These benchmark datasets are collected from the real world across different domains, including 
2 co-purchase networks, \emph{Amazon Computers}~\cite{shchur2019pitfallsgraphneuralnetwork} (or \emph{Amazon Com.} for short) with 13K nodes and 491K edges, and \emph{Amazon Photo}~\cite{shchur2019pitfallsgraphneuralnetwork} (or \emph{Amazon P.} for short), 3 citation networks, \emph{Cora}~\cite{mccallum2000automating}, \emph{Cora\_ML}~\cite{mccallum2000automating}, and \emph{Citeseer}~\cite{giles1998citeseer}, and 3 webpage datasets, \emph{Cornell}~\cite{WebKB}, \emph{Texas}~\cite{WebKB}, and \emph{Wisconsin}~\cite{WebKB}. }
\opt{online}{
\phead{Datasets.}
We conduct our experiments on 8 widely adopted benchmark datasets to evaluate the link prediction task. 
These benchmark datasets are collected from the real world across different domains, including 
2 co-purchase networks, \emph{Amazon Computers}~\cite{shchur2019pitfallsgraphneuralnetwork} (or \emph{Amazon Com.} for short) with 13K nodes and 491K edges, and \emph{Amazon Photo}~\cite{shchur2019pitfallsgraphneuralnetwork} (or \emph{Amazon P.} for short), 3 citation networks, \emph{Cora}~\cite{mccallum2000automating}, \emph{Cora\_ML}~\cite{mccallum2000automating}, and \emph{Citeseer}~\cite{giles1998citeseer}, and 3 webpage datasets, \emph{Cornell}~\cite{WebKB}, \emph{Texas}~\cite{WebKB}, and \emph{Wisconsin}~\cite{WebKB}. }

\opt{full}{
\begin{table}[t]
    \centering
    \caption{Statistics of the dataset}
    \begin{tabular}{@{}l|ccc@{}}
        \toprule
        Dataset & Nodes & Edges & Features \\ 
        \midrule
        \textit{ogbl-collab} & 235,868 & 1,285,465 & 128\\
        \textit{Cora} & 2,708 & 10,556 & 1,433 \\
        \textit{Citeseer} & 3,327 & 9,104 & 3,703 \\
        \textit{Cora\_ML} & 2,995 & 16,316 & 2,879 \\
        \textit{Cornell} & 183 & 227 & 1,703 \\
        \textit{Texas} & 183 & 279 & 1,703 \\
        \textit{Wisconsin} & 251 & 450 & 1,703 \\
        \textit{Amazon Photo} & 7,650 & 238,162 & 745 \\
        \textit{Amazon Computers} & 13,752 & 491,722 & 767 \\
        \bottomrule
    \end{tabular}
    \label{tab:dataset_statistics}
\end{table}
}

\phead{Training settings.}
We follow the previous experimental protocols~\cite{kipf2016, yang2023} for evaluating link prediction models. 
\opt{short}{The detailed settings can be found in the online full version of this paper~\cite{rep_material}.}
\opt{full}{The data split of train/validation/test is 85\%/5\%/10\%. We report the average performance and its standard deviation of 10 fixed random splits on each dataset. For the models, the backbone GNN encoder is APPNP~\cite{gasteiger2018combining}, and we employ Adam~\cite{kingma2017} as the optimizer. For the hyperparameters, the learning rate is set to $0.001$, weight decay is $5\times10^{-4}$, and dropout rate is $0.3$. The hidden dimension of the embedding is $256$, and the output dimension is $64$ for most datasets and would be adjusted for smaller datasets. With grid search, the parameters $\lambda_1$, $\lambda_2$, and $\lambda_3$ in Eq.~\ref{overall_loss} are set as $3$, $1$, and $3$, respectively. }
\opt{online}{The data split of train/validation/test is 85\%/5\%/10\%. We report the average performance and its standard deviation of 10 fixed random splits on each dataset. For the models, the backbone GNN encoder is APPNP~\cite{gasteiger2018combining}, and we employ Adam~\cite{kingma2017} as the optimizer. For the hyperparameters, the learning rate is set to $0.001$, weight decay is $5\times10^{-4}$, and dropout rate is $0.3$. The hidden dimension of the embedding is $256$, and the output dimension is $64$ for most datasets and would be adjusted for smaller datasets. With grid search, the parameters $\lambda_1$, $\lambda_2$, and $\lambda_3$ in Eq.~\ref{overall_loss} are set as $3$, $1$, and $3$, respectively. }
 
\phead{Baselines.}
We compare our proposed \texttt{CoEBA} with 10 state-of-the-art baselines, including 6 node embedding methods: i) \texttt{SECRET}~\cite{yang2023}, ii) \texttt{Neo-GNN}~\cite{yun2022}, iii) \texttt{NCNC}~\cite{wang2024}, iv) \texttt{S2GAE}~\cite{tan2022}, v) \texttt{VGNAE}~\cite{ahn2021}, and vi) \texttt{CIMAGE}~\cite{park2025cimage}; 4 subgraph-based methods, i.e., vii) \texttt{ML-Link}~\cite{zangari2024link}, viii) \texttt{PULL}~\cite{kim2025accurate}, ix) \texttt{SEAL} \cite{zhang2018}, and x) \texttt{BUDDY}~\cite{chamberlain2023}.
\opt{short}{The details of these baselines are presented in the online full version~\cite{rep_material}.}
\opt{full}{The details of these baselines are detailed in the following paragraph. }
\opt{online}{The details of these baselines are detailed in the following paragraph. }


\opt{full}{
For \texttt{SEAL}, \texttt{Neo-GNNs}, \texttt{BUDDY}, and \texttt{NCNC}, we use the code implementations provided by Li et al. \cite{Li2023evaluate}. For \texttt{VGNAE}, \texttt{S2GAE}, \texttt{SECRET}, \texttt{ML-Link}, \texttt{PULL}, and \texttt{CIMAGE}, we use the official implementations. 
The baseline settings for various models are summarized as follows: \texttt{SEAL}, \texttt{Neo-GNNs}, \texttt{VGNAE}, and \texttt{SECRET} use 64 hidden channels for both layers, \texttt{S2GAE} uses 256 channels for the first hidden layer and 64 for the second, \texttt{BUDDY} uses 256 channels for both hidden layers, and \texttt{NCNC} uses 32 channels for both. \texttt{SEAL} specifies a maximum of 25 nodes per hop and uses 1 hop\footnote{This configuration follows the implementation from Li et al.~\cite{Li2023evaluate}, and the resulting performance of \texttt{SEAL} aligns well with the results reported in Ma et al.~\cite{ma2024mixture}.}
, whereas the other models do not specify these parameters. \texttt{SEAL} and \texttt{Neo-GNNs} use 1 layer, \texttt{S2GAE} uses 2 layers, and \texttt{SEAL}, \texttt{Neo-GNNs}, and \texttt{BUDDY} use 2 layers for the predictor. All models are trained for 1000 epochs. The learning rate is set to 0.001 for \texttt{SEAL}, \texttt{Neo-GNNs}, \texttt{BUDDY}, \texttt{SECRET}, and \texttt{NCNC}, 0.005 for \texttt{VGNAE}, and 0.01 for \texttt{S2GAE}. The dropout rate is 0.5 for \texttt{S2GAE} and \texttt{BUDDY}, 0.3 for \texttt{SECRET}, and 0 for \texttt{SEAL}, with other models unspecified. Weight decay (L2) is set to 0 for \texttt{SEAL}, \texttt{Neo-GNNs}, and \texttt{BUDDY}, 0.0005 for \texttt{NCNC}, with others unspecified. \texttt{Neo-GNNs} use a feature edge dimension, feature node dimension, and G Phi dimension of 64. \texttt{S2GAE} uses a beta of 0.1 and a scaling factor of 1.8, with 2 decode layers and a mask ratio of 0.7. \texttt{NCNC} specifies additional parameters including XDP (0.4), TDP (0), PT (0.75), GNNEDP (0), PreEDP (0.55), GNNDP (0.75), probscale (6.5), proboffset (4.4), alpha (0.4), GNNLR (0.4), PreLR (0.0085), RMLayers (0.0078), gamma (1), delta (1), and uses maskinput, JK, use\_xlin, tailact, LN, and LNNN as TRUE. Parameters not listed are set to their default values as specified in the respective codebases.
}
\opt{online}{
For \texttt{SEAL}, \texttt{Neo-GNNs}, \texttt{BUDDY}, and \texttt{NCNC}, we use the code implementations provided by Li et al. \cite{Li2023evaluate}. For \texttt{VGNAE}, \texttt{S2GAE}, \texttt{SECRET}, \texttt{ML-Link}, \texttt{PULL}, and \texttt{CIMAGE}, we use the official implementations. 
The baseline settings for various models are summarized as follows: \texttt{SEAL}, \texttt{Neo-GNNs}, \texttt{VGNAE}, and \texttt{SECRET} use 64 hidden channels for both layers, \texttt{S2GAE} uses 256 channels for the first hidden layer and 64 for the second, \texttt{BUDDY} uses 256 channels for both hidden layers, and \texttt{NCNC} uses 32 channels for both. \texttt{SEAL} specifies a maximum of 25 nodes per hop and uses 1 hop\footnote{This configuration follows the implementation from Li et al.~\cite{Li2023evaluate}, and the resulting performance of \texttt{SEAL} aligns well with the results reported in Ma et al.~\cite{ma2024mixture}.}
, whereas the other models do not specify these parameters. \texttt{SEAL} and \texttt{Neo-GNNs} use 1 layer, \texttt{S2GAE} uses 2 layers, and \texttt{SEAL}, \texttt{Neo-GNNs}, and \texttt{BUDDY} use 2 layers for the predictor. All models are trained for 1000 epochs. The learning rate is set to 0.001 for \texttt{SEAL}, \texttt{Neo-GNNs}, \texttt{BUDDY}, \texttt{SECRET}, and \texttt{NCNC}, 0.005 for \texttt{VGNAE}, and 0.01 for \texttt{S2GAE}. The dropout rate is 0.5 for \texttt{S2GAE} and \texttt{BUDDY}, 0.3 for \texttt{SECRET}, and 0 for \texttt{SEAL}, with other models unspecified. Weight decay (L2) is set to 0 for \texttt{SEAL}, \texttt{Neo-GNNs}, and \texttt{BUDDY}, 0.0005 for \texttt{NCNC}, with others unspecified. \texttt{Neo-GNNs} use a feature edge dimension, feature node dimension, and G Phi dimension of 64. \texttt{S2GAE} uses a beta of 0.1 and a scaling factor of 1.8, with 2 decode layers and a mask ratio of 0.7. \texttt{NCNC} specifies additional parameters including XDP (0.4), TDP (0), PT (0.75), GNNEDP (0), PreEDP (0.55), GNNDP (0.75), probscale (6.5), proboffset (4.4), alpha (0.4), GNNLR (0.4), PreLR (0.0085), RMLayers (0.0078), gamma (1), delta (1), and uses maskinput, JK, use\_xlin, tailact, LN, and LNNN as TRUE. Parameters not listed are set to their default values as specified in the respective codebases.
}

\opt{full}{
\begin{table*}[t]
\caption{Hits@10(\%) with standard deviations on 9 benchmark datasets -- \textbf{Best performance: bold}, \underline{runner-up: underlined}; the average Pearson correlation coefficient between the minimum node degree and Hits@10 is 0.68}
\label{tab:result}
\resizebox{2.1\columnwidth}{!}{%
\def\arraystretch{1.2}
\tiny
\begin{tabular}{l|ccccccccccc}
\hline
 &
\begin{tabular}[c]{@{}c@{}} \textit{ogbl-collab}\end{tabular} &
\begin{tabular}[c]{@{}c@{}} \textit{Cora} \end{tabular} &
\begin{tabular}[c]{@{}c@{}} \textit{Citeseer}\end{tabular} &
\begin{tabular}[c]{@{}c@{}} \textit{Cora\_ML}\end{tabular} &
\begin{tabular}[c]{@{}c@{}} \textit{Cornell}\end{tabular} &
\begin{tabular}[c]{@{}c@{}} \textit{Texas}\end{tabular} &
\begin{tabular}[c]{@{}c@{}} \textit{Wisconsin}\end{tabular} &
\begin{tabular}[c]{@{}c@{}} \textit{Amazon Photo}\end{tabular} &
\begin{tabular}[c]{@{}c@{}} \textit{Amazon Com.}\end{tabular} 
\\\hline
\texttt{SEAL} (NIPS'18) & 35.46$\pm$6.23 & 51.31$\pm$5.13 & 62.75$\pm$0.89 & 47.70$\pm$3.25 & 75.30$\pm$7.75 & \underline{93.12$\pm$6.22} & 62.17$\pm$7.68 & 19.87$\pm$2.12 & 13.48$\pm$1.18\\ 
\texttt{Neo-GNNs} (NIPS'21) & 41.90$\pm$5.67 & 61.14$\pm$4.13 & 59.85$\pm$3.24 & 52.92$\pm$9.20 & 30.59$\pm$11.70 & 66.25$\pm$16.89 & 56.52$\pm$12.30 & 33.52$\pm$4.24 & \underline{14.63$\pm$3.75}\\ 
\texttt{VGNAE} (CIKM'21) & 44.92$\pm$3.63 & 76.83$\pm$0.74 & 73.45$\pm$1.21 & 65.70$\pm$1.06 & 54.37$\pm$5.62 & 56.87$\pm$8.12 & 66.08$\pm$6.67 & 25.01$\pm$1.96 & 14.03$\pm$0.22\\ 
\texttt{S2GAE} (WSDM'23) & 45.43$\pm$4.01 & 64.42$\pm$4.98 & 59.89$\pm$5.33 & 47.59$\pm$4.13 & 67.06$\pm$13.92 & 85.62$\pm$15.60 & 62.61$\pm$9.21 & 18.00$\pm$3.81 & 8.35$\pm$2.86\\  
\texttt{BUDDY} (ICLR'23) & 52.14$\pm$5.93 & 70.78$\pm$3.26 & 72.00$\pm$10.59 & 48.28$\pm$3.13 & 72.35$\pm$20.01 & 83.12$\pm$13.83 & \underline{82.61$\pm$12.30} & 34.85$\pm$2.70 & 13.77$\pm$2.19\\ 
\texttt{SECRET} (ECAI'23) & \underline{53.21$\pm$1.31} & \underline{78.00$\pm$0.45} & 74.46$\pm$0.35 & 65.84$\pm$0.82 & 70.58$\pm$00.00& 81.25$\pm$0.00 & 73.91$\pm$0.00 & 31.56$\pm$0.75 & 12.75$\pm$0.57\\  
\texttt{NCNC} (ICLR'24) & 49.38$\pm$3.14 & 65.37$\pm$3.64 & 67.78$\pm$2.37 & 63.32$\pm$8.95 & 54.70$\pm$17.98 & 44.38$\pm$18.27 & 42.17$\pm$13.29 & \underline{38.78$\pm$6.41} & 13.66$\pm$2.82\\
\texttt{ML-Link} (WWW'24) & 33.09$\pm$4.04 & 62.73$\pm$2.68 & 58.81$\pm$3.78 & 58.68$\pm$3.20 & 68.75$\pm$9.88 & 80.00$\pm$5.23 & 70.43$\pm$9.43 & 32.12$\pm$2.42 & 14.30$\pm$2.61\\
\texttt{PULL} (AAAI'25) & 34.07$\pm$2.56 & 62.81$\pm$1.52 & 56.09$\pm$1.06 & 70.28$\pm$1.73 & 71.25$\pm$3.42 & 80.00$\pm$2.80 & 53.91$\pm$4.96 & 22.09$\pm$1.78 & 13.84$\pm$0.80\\
\texttt{CIMAGE} (WSDM'25) & 36.54$\pm$3.34 & 70.17$\pm$4.89 & \underline{75.88$\pm$5.18} & \underline{70.92$\pm$2.53} & \underline{82.50$\pm$5.23} & 87.50$\pm$8.84 & 77.39$\pm$5.67 & 33.04$\pm$4.03 & 12.75$\pm$3.62\\ \hline
\texttt{CoEBA (ours)} & \bf{55.19$\pm$0.35} & \bf{79.54$\pm$0.36} & \bf{76.13$\pm$0.39} & \bf{71.06$\pm$1.00} & \bf{94.11$\pm$0.00} & \bf{93.75$\pm$0.00} & \bf{91.30$\pm$0.00} & \bf{38.92$\pm$0.48} & \bf{14.75$\pm$0.13}\\ \hline
\end{tabular}%
}
\end{table*}
}
\opt{short}{
\begin{table*}[t]
\caption{Hits@10(\%) with standard deviations on 8 benchmark datasets -- \textbf{Best performance: bold}, \underline{runner-up: underlined}; the average Pearson correlation coefficient between the minimum node degree and Hits@10 is 0.68}
\label{tab:result}
\centering
\setlength{\tabcolsep}{1.4\tabcolsep}
\small
\begin{tabular}{l|ccccccccccc}
\hline
 &
\begin{tabular}[c]{@{}c@{}} \textit{Cora} \end{tabular} &
\begin{tabular}[c]{@{}c@{}} \textit{Citeseer}\end{tabular} &
\begin{tabular}[c]{@{}c@{}} \textit{Cora\_ML}\end{tabular} &
\begin{tabular}[c]{@{}c@{}} \textit{Cornell}\end{tabular} &
\begin{tabular}[c]{@{}c@{}} \textit{Texas}\end{tabular} &
\begin{tabular}[c]{@{}c@{}} \textit{Wisconsin}\end{tabular} &
\begin{tabular}[c]{@{}c@{}} \textit{Amazon Photo}\end{tabular} &
\begin{tabular}[c]{@{}c@{}} \textit{Amazon Com.}\end{tabular} 
\\\hline
\texttt{SEAL} (NIPS'18)  & 51.31$\pm$5.13 & 62.75$\pm$0.89 & 47.70$\pm$3.25 & 75.30$\pm$7.75 & \underline{93.12$\pm$6.22} & 62.17$\pm$7.68 & 19.87$\pm$2.12 & 13.48$\pm$1.18\\ 
\texttt{Neo-GNNs} (NIPS'21) & 61.14$\pm$4.13 & 59.85$\pm$3.24 & 52.92$\pm$9.20 & 30.59$\pm$11.70 & 66.25$\pm$16.89 & 56.52$\pm$12.30 & 33.52$\pm$4.24 & \underline{14.63$\pm$3.75}\\ 
\texttt{VGNAE} (CIKM'21) & 76.83$\pm$0.74 & 73.45$\pm$1.21 & 65.70$\pm$1.06 & 54.37$\pm$5.62 & 56.87$\pm$8.12 & 66.08$\pm$6.67 & 25.01$\pm$1.96 & 14.03$\pm$0.22\\ 
\texttt{S2GAE} (WSDM'23)  & 64.42$\pm$4.98 & 59.89$\pm$5.33 & 47.59$\pm$4.13 & 67.06$\pm$13.92 & 85.62$\pm$15.60 & 62.61$\pm$9.21 & 18.00$\pm$3.81 & 8.35$\pm$2.86\\  
\texttt{BUDDY} (ICLR'23) & 70.78$\pm$3.26 & 72.00$\pm$10.59 & 48.28$\pm$3.13 & 72.35$\pm$20.01 & 83.12$\pm$13.83 & \underline{82.61$\pm$12.30} & 34.85$\pm$2.70 & 13.77$\pm$2.19\\ 
\texttt{SECRET} (ECAI'23)  & \underline{78.00$\pm$0.45} & 74.46$\pm$0.35 & 65.84$\pm$0.82 & 70.58$\pm$00.00& 81.25$\pm$0.00 & 73.91$\pm$0.00 & 31.56$\pm$0.75 & 12.75$\pm$0.57\\  
\texttt{NCNC} (ICLR'24) & 65.37$\pm$3.64 & 67.78$\pm$2.37 & 63.32$\pm$8.95 & 54.70$\pm$17.98 & 44.38$\pm$18.27 & 42.17$\pm$13.29 & \underline{38.78$\pm$6.41} & 13.66$\pm$2.82\\
\texttt{ML-Link} (WWW'24)  & 62.73$\pm$2.68 & 58.81$\pm$3.78 & 58.68$\pm$3.20 & 68.75$\pm$9.88 & 80.00$\pm$5.23 & 70.43$\pm$9.43 & 32.12$\pm$2.42 & 14.30$\pm$2.61\\
\texttt{PULL} (AAAI'25)  & 62.81$\pm$1.52 & 56.09$\pm$1.06 & 70.28$\pm$1.73 & 71.25$\pm$3.42 & 80.00$\pm$2.80 & 53.91$\pm$4.96 & 22.09$\pm$1.78 & 13.84$\pm$0.80\\
\texttt{CIMAGE} (WSDM'25)  & 70.17$\pm$4.89 & \underline{75.88$\pm$5.18} & \underline{70.92$\pm$2.53} & \underline{82.50$\pm$5.23} & 87.50$\pm$8.84 & 77.39$\pm$5.67 & 33.04$\pm$4.03 & 12.75$\pm$3.62\\ \hline
\texttt{CoEBA (ours)} & \bf{79.54$\pm$0.36} & \bf{76.13$\pm$0.39} & \bf{71.06$\pm$1.00} & \bf{94.11$\pm$0.00} & \bf{93.75$\pm$0.00} & \bf{91.30$\pm$0.00} & \bf{38.92$\pm$0.48} & \bf{14.75$\pm$0.13}\\ \hline
\end{tabular}%

\end{table*}
}
\opt{online}{
\begin{table*}[t]
\caption{Hits@10(\%) with standard deviations on 8 benchmark datasets -- \textbf{Best performance: bold}, \underline{runner-up: underlined}; the average Pearson correlation coefficient between the minimum node degree and Hits@10 is 0.68}
\label{tab:result}
\centering
\setlength{\tabcolsep}{1.4\tabcolsep}
\small
\begin{tabular}{l|ccccccccccc}
\hline
 &
\begin{tabular}[c]{@{}c@{}} \textit{Cora} \end{tabular} &
\begin{tabular}[c]{@{}c@{}} \textit{Citeseer}\end{tabular} &
\begin{tabular}[c]{@{}c@{}} \textit{Cora\_ML}\end{tabular} &
\begin{tabular}[c]{@{}c@{}} \textit{Cornell}\end{tabular} &
\begin{tabular}[c]{@{}c@{}} \textit{Texas}\end{tabular} &
\begin{tabular}[c]{@{}c@{}} \textit{Wisconsin}\end{tabular} &
\begin{tabular}[c]{@{}c@{}} \textit{Amazon Photo}\end{tabular} &
\begin{tabular}[c]{@{}c@{}} \textit{Amazon Com.}\end{tabular} 
\\\hline
\texttt{SEAL} (NIPS'18)  & 51.31$\pm$5.13 & 62.75$\pm$0.89 & 47.70$\pm$3.25 & 75.30$\pm$7.75 & \underline{93.12$\pm$6.22} & 62.17$\pm$7.68 & 19.87$\pm$2.12 & 13.48$\pm$1.18\\ 
\texttt{Neo-GNNs} (NIPS'21) & 61.14$\pm$4.13 & 59.85$\pm$3.24 & 52.92$\pm$9.20 & 30.59$\pm$11.70 & 66.25$\pm$16.89 & 56.52$\pm$12.30 & 33.52$\pm$4.24 & \underline{14.63$\pm$3.75}\\ 
\texttt{VGNAE} (CIKM'21) & 76.83$\pm$0.74 & 73.45$\pm$1.21 & 65.70$\pm$1.06 & 54.37$\pm$5.62 & 56.87$\pm$8.12 & 66.08$\pm$6.67 & 25.01$\pm$1.96 & 14.03$\pm$0.22\\ 
\texttt{S2GAE} (WSDM'23)  & 64.42$\pm$4.98 & 59.89$\pm$5.33 & 47.59$\pm$4.13 & 67.06$\pm$13.92 & 85.62$\pm$15.60 & 62.61$\pm$9.21 & 18.00$\pm$3.81 & 8.35$\pm$2.86\\  
\texttt{BUDDY} (ICLR'23) & 70.78$\pm$3.26 & 72.00$\pm$10.59 & 48.28$\pm$3.13 & 72.35$\pm$20.01 & 83.12$\pm$13.83 & \underline{82.61$\pm$12.30} & 34.85$\pm$2.70 & 13.77$\pm$2.19\\ 
\texttt{SECRET} (ECAI'23)  & \underline{78.00$\pm$0.45} & 74.46$\pm$0.35 & 65.84$\pm$0.82 & 70.58$\pm$00.00& 81.25$\pm$0.00 & 73.91$\pm$0.00 & 31.56$\pm$0.75 & 12.75$\pm$0.57\\  
\texttt{NCNC} (ICLR'24) & 65.37$\pm$3.64 & 67.78$\pm$2.37 & 63.32$\pm$8.95 & 54.70$\pm$17.98 & 44.38$\pm$18.27 & 42.17$\pm$13.29 & \underline{38.78$\pm$6.41} & 13.66$\pm$2.82\\
\texttt{ML-Link} (WWW'24)  & 62.73$\pm$2.68 & 58.81$\pm$3.78 & 58.68$\pm$3.20 & 68.75$\pm$9.88 & 80.00$\pm$5.23 & 70.43$\pm$9.43 & 32.12$\pm$2.42 & 14.30$\pm$2.61\\
\texttt{PULL} (AAAI'25)  & 62.81$\pm$1.52 & 56.09$\pm$1.06 & 70.28$\pm$1.73 & 71.25$\pm$3.42 & 80.00$\pm$2.80 & 53.91$\pm$4.96 & 22.09$\pm$1.78 & 13.84$\pm$0.80\\
\texttt{CIMAGE} (WSDM'25)  & 70.17$\pm$4.89 & \underline{75.88$\pm$5.18} & \underline{70.92$\pm$2.53} & \underline{82.50$\pm$5.23} & 87.50$\pm$8.84 & 77.39$\pm$5.67 & 33.04$\pm$4.03 & 12.75$\pm$3.62\\ \hline
\texttt{CoEBA (ours)} & \bf{79.54$\pm$0.36} & \bf{76.13$\pm$0.39} & \bf{71.06$\pm$1.00} & \bf{94.11$\pm$0.00} & \bf{93.75$\pm$0.00} & \bf{91.30$\pm$0.00} & \bf{38.92$\pm$0.48} & \bf{14.75$\pm$0.13}\\ \hline
\end{tabular}%

\end{table*}
}

\phead{Evaluation metric.}
As reported in previous work~\cite{Li2023evaluate, huang2023linkpredictiongraphneural, Yang2015}, AUC (Area Under Curve) may not be a proper metric for link prediction. Thus, following recent works~\cite{wang2024,zhao2022,chamberlain2023}, we adopt \emph{Hits@10} as the performance metric. Higher Hits@10 indicates better performance. Standard deviations are also reported to quantify variability.

\phead{Reproducibility materials.} The reproducibility materials, such as codes, models, and documents, are available upon request.

\phead{Performance comparisons.}
Table~\ref{tab:result} presents the Hits@10 of all the baseline models and our proposed method. The results demonstrate that our proposed \texttt{CoEBA} achieves the best Hits@10 for most datasets. Although \texttt{VGNAE}, \texttt{SECRET}, and the proposed \texttt{CoEBA} all employ \texttt{VGNAE} as the backbone autoencoder, we significantly enhance the performance by leveraging our proposed \emph{Edge Balancing Augmentation (EBA)}  and contrastive losses. On the other hand, \texttt{S2GAE} exhibits inferior performance across almost all datasets compared to the other state-of-the-art methods. 

Although the subgraph-based methods, i.e., \texttt{SEAL} and \texttt{BUDDY}, exhibit a higher expressive ability, our proposed \texttt{CoEBA} still outperforms them because the minimum node degree is increased to ensure more concentrated embeddings generated by \texttt{CoEBA}.
To further confirm the importance of this design, we examine the relationship between the minimum node degree and the model's link prediction performance. The results reveal a strong positive correlation between the minimum node degree and Hits@10, with Pearson correlation coefficients of $0.83$ on \textit{Cora} and $0.69$ on \textit{Citeseer}. This supports our theoretical insight from Corollary~\ref{cor:direction} that increasing the minimum node degree improves latent embedding concentration, which in turn enhances generalization in link prediction. 


In contrast, the subgraph-based methods do not take this into consideration, leading to much inferior performance. Furthermore, both \texttt{Neo-GNNs} and \texttt{NCNC} take neighborhood relationship into consideration, resulting in good performance, particularly in large datasets like \emph{Amazon Photo}, \emph{Amazon Computer}, and \emph{LastFMAsia}. However, our \texttt{CoEBA} still significantly outperforms them in all the datasets, except that \texttt{Neo-GNNs} achieves slightly better performance on \emph{Amazon Computer}. This is because \texttt{Neo-GNNs} performs the time-consuming multi-hop neighbor aggregation. However, \texttt{Neo-GNNs} requires significantly more time to complete its task. Please note that \texttt{Neo-GNNs} is significantly outperformed by our approach with $16\%-63\%$ Hits@10 in other datasets, such as \emph{Cora}, \emph{Citeseer}, \emph{Cora\_ML}, \emph{Texas}, and \emph{Wisconsin}. 
Also, our \texttt{CoEBA} exhibits a much smaller standard deviation in performance across all datasets, suggesting greater stability compared to other state-of-the-art models, indicating that \texttt{CoEBA} is more reliable and consistent in its performance.

\opt{full}{
\begin{table}[t]
    \centering
    \scriptsize  
    \caption{Hits@10(\%) under (a) HeaRT~\cite{Li2023evaluate} setting and (b) the ablation study of \texttt{CoEBA}}
    \hspace{-4em}
    \begin{subtable}[t]{0.45\columnwidth}
        \caption{Evaluation under HeaRT}
        \label{tab:heart_results}
        \renewcommand{\arraystretch}{1.1}
        \begin{tabular}{l|cc}
            \hline
            &  \textit{Cora}  & \textit{Citeseer} \\
            \hline
            \texttt{SEAL}         & 24.27 & 42.37 \\
            \texttt{Neo-GNN}      & 29.27 & 41.74 \\
            \texttt{BUDDY}        & 36.70 & 53.72 \\
            \texttt{SECRET}       & 38.71 & 54.07 \\
            \texttt{NCNC}         & 30.40 & 48.35 \\ \hline
            \texttt{CoEBA (ours)} & \textbf{42.31} & \textbf{57.58} \\
            \hline
        \end{tabular}
    \end{subtable}
    \begin{subtable}[t]{0.45\columnwidth}
        \caption{Ablation study}
        \label{tab:ablation}
        \renewcommand{\arraystretch}{1.1}
        \begin{tabular}{l|cc}
            \hline
            Model Config. & \textit{Cora} & \textit{Citeseer} \\
            \hline
            \texttt{w/o EBA \& CL} & 76.66 $\pm$ 0.00 & 68.96 $\pm$ 0.09 \\
            \texttt{w/o CL}        & 77.53 $\pm$ 0.25 & 71.51 $\pm$ 0.39 \\
            \texttt{w/o within-CL} & 77.72 $\pm$ 0.28 & 73.40 $\pm$ 0.41 \\
            \texttt{w/o btn-CL}    & 79.08 $\pm$ 0.52 & 72.26 $\pm$ 0.47 \\
            \texttt{w/o EBA}       & 75.78 $\pm$ 0.10 & 71.12 $\pm$ 0.12 \\ \hline
            \texttt{CoEBA}         & \textbf{79.54} $\pm$ 0.36 & \textbf{76.13} $\pm$ 0.39 \\
            \hline
        \end{tabular}
    \end{subtable}
\end{table}
}
\opt{short}{
\begin{table}[t]{}
        \caption{The ablation study of \texttt{CoEBA}}
        \label{tab:ablation}
        \setlength{\tabcolsep}{1.4\tabcolsep}
        \small
        \begin{tabular}{l|cc}
            \hline
            Model Config. & \textit{Cora} & \textit{Citeseer} \\
            \hline
            \texttt{w/o EBA \& CL} & 76.66 $\pm$ 0.00 & 68.96 $\pm$ 0.09 \\
            \texttt{w/o CL}        & 77.53 $\pm$ 0.25 & 71.51 $\pm$ 0.39 \\
            \texttt{w/o within-CL} & 77.72 $\pm$ 0.28 & 73.40 $\pm$ 0.41 \\
            \texttt{w/o btn-CL}    & 79.08 $\pm$ 0.52 & 72.26 $\pm$ 0.47 \\
            \texttt{w/o EBA}       & 75.78 $\pm$ 0.10 & 71.12 $\pm$ 0.12 \\ \hline
            \texttt{CoEBA}         & \textbf{79.54} $\pm$ 0.36 & \textbf{76.13} $\pm$ 0.39 \\
            \hline
        \end{tabular}
    \end{table}
}
\opt{online}{
\begin{table}[t]{}
        \caption{The ablation study of \texttt{CoEBA}}
        \label{tab:ablation}
        \setlength{\tabcolsep}{1.4\tabcolsep}
        \small
        \begin{tabular}{l|cc}
            \hline
            Model Config. & \textit{Cora} & \textit{Citeseer} \\
            \hline
            \texttt{w/o EBA \& CL} & 76.66 $\pm$ 0.00 & 68.96 $\pm$ 0.09 \\
            \texttt{w/o CL}        & 77.53 $\pm$ 0.25 & 71.51 $\pm$ 0.39 \\
            \texttt{w/o within-CL} & 77.72 $\pm$ 0.28 & 73.40 $\pm$ 0.41 \\
            \texttt{w/o btn-CL}    & 79.08 $\pm$ 0.52 & 72.26 $\pm$ 0.47 \\
            \texttt{w/o EBA}       & 75.78 $\pm$ 0.10 & 71.12 $\pm$ 0.12 \\ \hline
            \texttt{CoEBA}         & \textbf{79.54} $\pm$ 0.36 & \textbf{76.13} $\pm$ 0.39 \\
            \hline
        \end{tabular}
    \end{table}
}

\opt{full}{
\phead{Performance under a more challenging HeaRT~\cite{Li2023evaluate} setting. }In addition, we also compare \texttt{CoEBA} with baselines under the HeaRT~\cite{Li2023evaluate} setting, which is a sampling strategy to create more challenging testing data. We generate the testing data following HeaRT and obtain the Hits@10 for \texttt{CoEBA} and baselines. Table~\ref{tab:heart_results} shows the results of \emph{Cora} and \emph{Citeseer}. 
Specifically, in \textit{Cora}, Hits@10 of \texttt{CoEBA} is 42.31\%, and Hits@10 of the runner-up, \texttt{SECRET}, is 38.71\%. In \textit{Citeseer}, Hits@10 of \texttt{CoEBA} is 57.58\%, and Hits@10 of the runner-up, \texttt{SECRET}, is 54.07\%. 
}

\phead{Ablation study.}
To demonstrate the effectiveness of each module in our \texttt{CoEBA}, we conduct ablation studies on two datasets: \emph{Cora}, and \emph{Citeseer}. The results are summarized in Table \ref{tab:ablation}. Here, \texttt{w/o EBA} indicates the proposed \texttt{CoEBA} without \emph{Edge Balancing Augmentation (EBA)}, i.e., employing the original graph as the augmented graph. The terms \texttt{w/o btn-CL} and \texttt{w/o within-CL} denote the removal of the two losses, respectively, while \texttt{w/o CL} indicates all the contrastive losses are excluded. Finally, \texttt{w/o EBA \& CL} indicates neither EBA nor any contrastive loss is incorporated in \texttt{CoEBA}.

Table~\ref{tab:ablation} indicates that each component effectively enhances model performance. The poor performance of \texttt{w/o EBA} and \texttt{w/o EBA \& CL} indicates that the proposed augmentation approach, \emph{EBA}, and the proposed contrastive losses play critical roles in the overall performance. In addition, \texttt{w/o CL} also indicates the importance of the contrastive losses. The proposed approach, \texttt{CoEBA}, which is equipped with all the components, significantly outperforms all the other baselines. This indicates that integrating all the components into our proposed approach significantly boosts overall performance.

\begin{table}[t]
    \centering
    \caption{Hits@10(\%) for various autoencoder-based models before and after equipped with EBA on various datasets}
    \label{tab:plug-and-play}
    \setlength{\tabcolsep}{1.4\tabcolsep}
    \small
    \begin{tabular}{l|cc}
        \hline
        Model & \textit{Cora} & \textit{Citeseer} \\
        \hline
        \texttt{GAE}          & $65.24 \pm 1.35$ & $55.41 \pm 1.49$ \\
        \texttt{GAE + EBA}    & $66.91 \pm 1.97$ & $66.81 \pm 3.53$ \\
        \texttt{GNAE}         & $76.85 \pm 0.94$ & $73.25 \pm 2.69$ \\
        \texttt{GNAE + EBA}   & $76.47 \pm 1.59$ & $77.78 \pm 1.43$ \\
        \texttt{VGNAE}        & $76.83 \pm 0.74$ & $73.45 \pm 1.21$ \\
        \texttt{VGNAE + EBA}  & $79.87 \pm 0.67$ & $77.54 \pm 2.05$ \\
        \hline
    \end{tabular}
\end{table}

\phead{Integrating EBA with different autoencoder-based approaches.}
As stated earlier, our proposed EBA can act as a plug-and-play module to be integrated with existing autoencoder-based approaches that utilize the reconstruction loss to enhance model performance. 
In Table~\ref{tab:plug-and-play}, we demonstrate the results of equipping EBA with the existing \texttt{GAE}~\cite{kipf2016}, \texttt{GNAE}~\cite{ahn2021}, and \texttt{VGNAE}~\cite{ahn2021} on the \emph{Cora} and \emph{Citeseer} datasets. Here, we utilize the latent embeddings learned by their models for EBA. We also perform the feature augmentation and employ the same neighbor-concentrated contrastive losses.

Compared to their original versions, the integration of EBA significantly improves their performance. This validates the effectiveness of the proposed EBA, indicating that it is indeed beneficial and can function as a plug-and-play module to existing approaches. It is worth noting that, although the variance increases slightly after equipping \texttt{CoEBA}, it remains minimal, as compared to other baseline approaches. As shown in Table~\ref{tab:result}, the variance of \texttt{CoEBA} across all datasets is less than $1$, but the variance of other baselines may have variances reaching up to $20$. 
Furthermore, while our experiments focus on autoencoder-based models, the design of EBA is model-agnostic and can be easily adapted to other graph representation learning frameworks with minor adjustments. 

\opt{full}{
\begin{table}[t]
    \centering
    \caption{Efficiency on $4$ datasets. Each element shows \textit{training time per epoch (sec)} / \textit{inference time per test case (sec)}}
    \label{tab:efficiency}
    \resizebox{.9\columnwidth}{!}{
    \begin{tabular}{l|cccc}
        \hline
        & \textit{ogbl-collab} & \textit{Cora} & \textit{Citeseer} & \textit{Amazon Com.} \\
        \hline
        \texttt{SEAL} & 49.18/14.03 & 0.47/0.10 & 0.55/0.13 & 37.74/10.02\\
        \texttt{Neo-GNN} & 948.36/13.89 & 0.43/0.03 & 0.31/0.02 & 775.14/10.69\\
        \texttt{BUDDY} & 62.48/10.89 & 0.05/0.03 & 0.07/0.04 & 1.60/7.19\\
        \texttt{SECRET} & 39.81/0.019 & 0.05/0.001 & 0.09/0.001 & 1.21/0.013\\
        \texttt{NCNC} & 18.13/5.01 & 0.87/0.07 & 1.05/0.07 & 14.37/3.74\\ \hline
        \texttt{CoEBA (ours)} & 47.16/0.018 & 0.07/0.001 & 0.10/0.001 & 1.27/0.016\\
        \hline
    \end{tabular}
    }
\end{table}
}
\opt{short}{
\begin{table}[t]
    \centering
    \caption{Efficiency on $4$ datasets. Each element shows \textit{training time per epoch (sec)} / \textit{inference time per test case (sec)}}
    \label{tab:efficiency}
    \setlength{\tabcolsep}{1.4\tabcolsep}
    \small
    \begin{tabular}{l|ccc}
        \hline
         & \textit{Cora} & \textit{Citeseer} & \textit{Amazon Com.} \\
        \hline
        \texttt{SEAL}  & 0.47/0.10 & 0.55/0.13 & 37.74/10.02\\
        \texttt{Neo-GNN}  & 0.43/0.03 & 0.31/0.02 & 775.14/10.69\\
        \texttt{BUDDY}  & 0.05/0.03 & 0.07/0.04 & 1.60/7.19\\
        \texttt{SECRET}  & 0.05/0.001 & 0.09/0.001 & 1.21/0.013\\
        \texttt{NCNC}  & 0.87/0.07 & 1.05/0.07 & 14.37/3.74\\ \hline
        \texttt{CoEBA (ours)}  & 0.07/0.001 & 0.10/0.001 & 1.27/0.016\\
        \hline
    \end{tabular}
    
\end{table}
}
\opt{online}{
\begin{table}[t]
    \centering
    \caption{Efficiency on $4$ datasets. Each element shows \textit{training time per epoch (sec)} / \textit{inference time per test case (sec)}}
    \label{tab:efficiency}
    \setlength{\tabcolsep}{1.4\tabcolsep}
    \small
    \begin{tabular}{l|ccc}
        \hline
         & \textit{Cora} & \textit{Citeseer} & \textit{Amazon Com.} \\
        \hline
        \texttt{SEAL}  & 0.47/0.10 & 0.55/0.13 & 37.74/10.02\\
        \texttt{Neo-GNN}  & 0.43/0.03 & 0.31/0.02 & 775.14/10.69\\
        \texttt{BUDDY}  & 0.05/0.03 & 0.07/0.04 & 1.60/7.19\\
        \texttt{SECRET}  & 0.05/0.001 & 0.09/0.001 & 1.21/0.013\\
        \texttt{NCNC}  & 0.87/0.07 & 1.05/0.07 & 14.37/3.74\\ \hline
        \texttt{CoEBA (ours)}  & 0.07/0.001 & 0.10/0.001 & 1.27/0.016\\
        \hline
    \end{tabular}
    
\end{table}
}

\phead{Efficiency.}
We compare the computational efficiency of the proposed \texttt{CoEBA} with other baselines in terms of \textit{training time per epoch} and \textit{inference time per test case}, as presented in Table~\ref{tab:efficiency}. 
\opt{full}{
On larger datasets like \textit{ogbl-collab} and \textit{Amazon Computers}, we observe that \texttt{CoEBA} outperforms most baselines and is only slightly behind \texttt{SECRET}. 
}
\opt{short}{
On a larger dataset like \textit{Amazon Computers}, we observe that \texttt{CoEBA} outperforms most baselines and is only slightly behind \texttt{SECRET}. 
}
\opt{online}{
On a larger dataset like \textit{Amazon Computers}, we observe that \texttt{CoEBA} outperforms most baselines and is only slightly behind \texttt{SECRET}. 
}
For smaller datasets, i.e., \textit{Cora} and \textit{Citeseer}, \texttt{CoEBA} provides competitive efficiency. This is because the baselines, such as \texttt{SEAL} and \texttt{BUDDY}, are subgraph-based methods, which require extracting multiple subgraphs, making them less efficient on large datasets. 

It is worth noting that although the complexity of our proposed method \texttt{CoEBA} is $O(N^2 \times D_f)$, it remains efficient in practice. Specifically, on smaller datasets such as \textit{Cora} and \textit{Citeseer}, \texttt{CoEBA} completes training in less than $0.15$ seconds per epoch and performs inference in under $1$ millisecond. 
\opt{full}{
On larger datasets like \textit{ogbl-collab} and \textit{Amazon Computers}, the training time remains under $50$ seconds and inference time below $20$ milliseconds per test case. 
}
\opt{short}{
On a larger dataset like  \textit{Amazon Computers}, the training time remains under $50$ seconds and inference time below $20$ milliseconds per test case. 
}
\opt{online}{
On a larger dataset like  \textit{Amazon Computers}, the training time remains under $50$ seconds and inference time below $20$ milliseconds per test case. 
}


\opt{full}{
\begin{figure}[t]
    \centering
    \includegraphics[width=0.9\columnwidth]{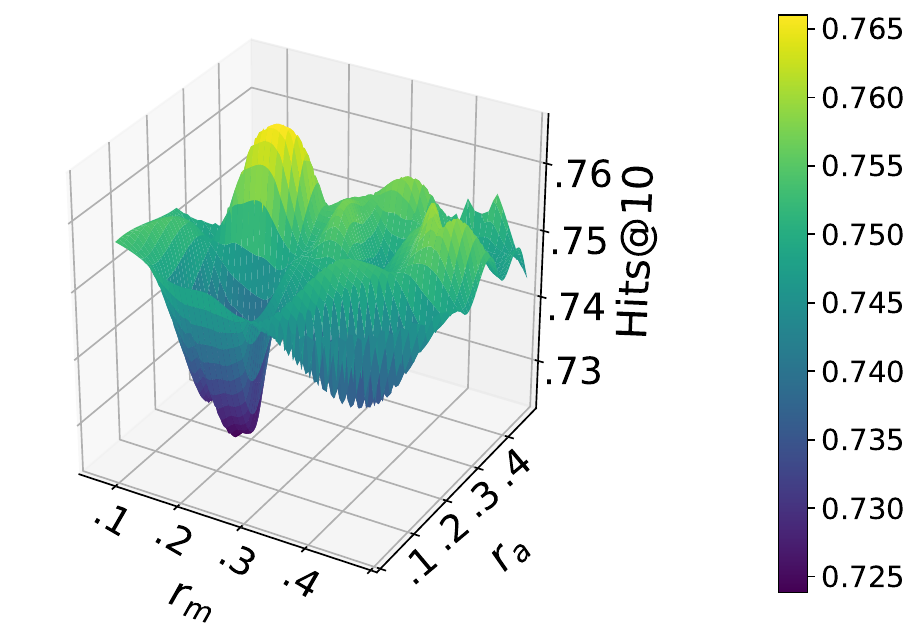}
    \caption{Sensitivity tests of $r_m$ and $r_a$ on dataset \emph{Cora}}
    \label{fig:sensitivity}
\end{figure}
}

\opt{online}{
\begin{figure}[t]
    \centering
    \includegraphics[width=0.9\columnwidth]{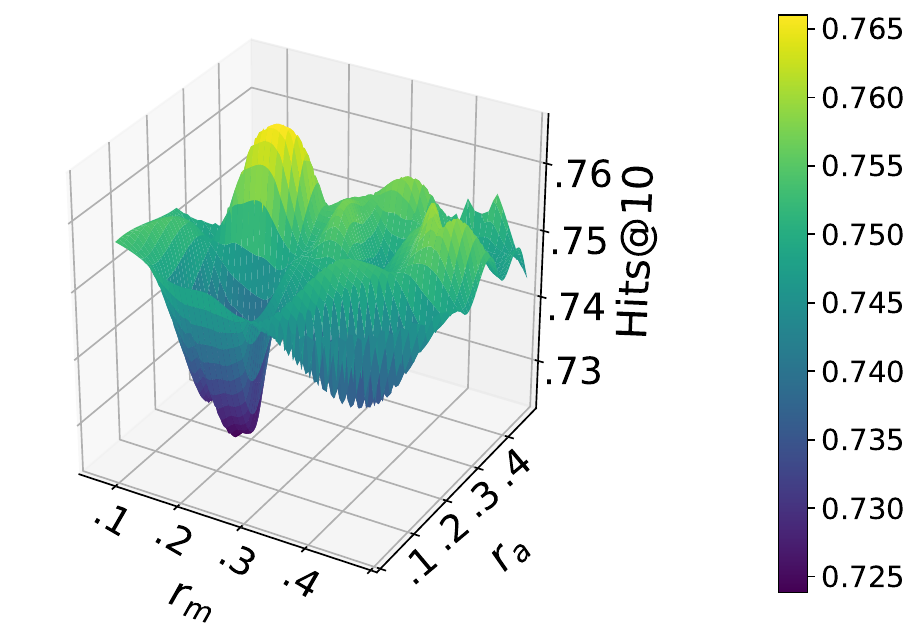}
    \caption{Sensitivity tests of $r_m$ and $r_a$ on dataset \emph{Cora}}
    \label{fig:sensitivity}
\end{figure}
}

\opt{short}{
\phead{Sensitivity test on $r_a$ and $r_m$.}
We conduct the sensitivity tests on dataset \emph{Cora}. The detailed experimental results are provided in the online full version~\cite{rep_material}. Based on the results, we set $r_m$ as 0.14 and $r_a$ as 0.40 for our experiments.}
\opt{full}{
\phead{Sensitivity test on $r_a$ and $r_m$.}
Fig.~\ref{fig:sensitivity} presents the sensitivity tests on the dataset \emph{Cora}. Here, we aim to explore the impact of varying the  \emph{neighbor removal ratio} $r_m$ and the \emph{neighbor addition ratio $r_a$}. 
To better illustrate the trade-off between these parameters, we exclude the feature augmentation.
As shown in the figure, the ratios impact the model performance. The performance peaks when the neighbor removal ratio is low and the neighbor addition ratio is high. However, when these parameters deviate from certain ranges, the performance declines, suggesting that these parameters significantly impact model performance. Based on the results, we set $r_m$ as 0.14 and $r_a$ as 0.40 for our experiments in this paper.
}
\opt{online}{
\phead{Sensitivity test on $r_a$ and $r_m$.}
Fig.~\ref{fig:sensitivity} presents the sensitivity tests on the dataset \emph{Cora}. Here, we aim to explore the impact of varying the  \emph{neighbor removal ratio} $r_m$ and the \emph{neighbor addition ratio $r_a$}. 
To better illustrate the trade-off between these parameters, we exclude the feature augmentation.
As shown in the figure, the ratios impact the model performance. The performance peaks when the neighbor removal ratio is low and the neighbor addition ratio is high. However, when these parameters deviate from certain ranges, the performance declines, suggesting that these parameters significantly impact model performance. Based on the results, we set $r_m$ as 0.14 and $r_a$ as 0.40 for our experiments in this paper.
}

\vspace{-5pt}
\section{Conclusion}
\label{sec:conclusion}
In this paper, we provide the first theoretical analysis of contrastive learning on link prediction and propose our new augmentation strategy, \emph{EBA}, which adjusts the node degrees in the graph. We also propose our approach, \texttt{CoEBA}, that integrates the proposed EBA and the new contrastive losses. Experimental results on multiple benchmark datasets demonstrate that our approach significantly outperforms the other state-of-the-art baseline models.



\section*{Acknowledgement}
This research was supported in part by the National Science and Technology Council (NSTC), Taiwan, under Grant NSTC 113-2628-E-007-012-MY3, NSTC 112-2221-E-007-088-MY3, and NSTC 114-2222-E-008-002-MY2. The authors sincerely appreciate this financial support, which was essential to the successful completion of the study.

\section*{GenAI Disclosure Statement}
The authors confirm that no generative AI tools are used in the writing or preparation of this work.

\bibliographystyle{ACM-Reference-Format}
\balance
\bibliography{reference}


\begin{thebibliography}{50}


\ifx \showCODEN    \undefined \def \showCODEN     #1{\unskip}     \fi
\ifx \showDOI      \undefined \def \showDOI       #1{#1}\fi
\ifx \showISBNx    \undefined \def \showISBNx     #1{\unskip}     \fi
\ifx \showISBNxiii \undefined \def \showISBNxiii  #1{\unskip}     \fi
\ifx \showISSN     \undefined \def \showISSN      #1{\unskip}     \fi
\ifx \showLCCN     \undefined \def \showLCCN      #1{\unskip}     \fi
\ifx \shownote     \undefined \def \shownote      #1{#1}          \fi
\ifx \showarticletitle \undefined \def \showarticletitle #1{#1}   \fi
\ifx \showURL      \undefined \def \showURL       {\relax}        \fi
\providecommand\bibfield[2]{#2}
\providecommand\bibinfo[2]{#2}
\providecommand\natexlab[1]{#1}
\providecommand\showeprint[2][]{arXiv:#2}

\bibitem[Ahn and Kim(2021)]%
        {ahn2021}
\bibfield{author}{\bibinfo{person}{Seong~Jin Ahn} {and} \bibinfo{person}{Myoung~Ho Kim}.} \bibinfo{year}{2021}\natexlab{}.
\newblock \showarticletitle{Variational Graph Normalized Autoencoders}. In \bibinfo{booktitle}{\emph{CIKM}}.
\newblock


\bibitem[{Carnegie Mellon University}(2001)]%
        {WebKB}
\bibfield{author}{\bibinfo{person}{{Carnegie Mellon University}}.} \bibinfo{year}{2001}\natexlab{}.
\newblock \bibinfo{title}{WebKB Project}.
\newblock \bibinfo{howpublished}{\url{https://www.cs.cmu.edu/project/theo-11/www/wwkb/}}.
\newblock


\bibitem[Chamberlain et~al\mbox{.}(2023)]%
        {chamberlain2023}
\bibfield{author}{\bibinfo{person}{Benjamin~Paul Chamberlain}, \bibinfo{person}{Sergey Shirobokov}, \bibinfo{person}{Emanuele Rossi}, \bibinfo{person}{Fabrizio Frasca}, \bibinfo{person}{Thomas Markovich}, \bibinfo{person}{Nils Hammerla}, \bibinfo{person}{Michael~M. Bronstein}, {and} \bibinfo{person}{Max Hansmire}.} \bibinfo{year}{2023}\natexlab{}.
\newblock \showarticletitle{Graph Neural Networks for Link Prediction with Subgraph Sketching}. In \bibinfo{booktitle}{\emph{ICLR}}.
\newblock


\bibitem[Chen and Kou(2023)]%
        {Chen_Kou_2023}
\bibfield{author}{\bibinfo{person}{Jialu Chen} {and} \bibinfo{person}{Gang Kou}.} \bibinfo{year}{2023}\natexlab{}.
\newblock \showarticletitle{Attribute and structure preserving graph contrastive learning}. In \bibinfo{booktitle}{\emph{AAAI}}. \bibinfo{pages}{7024--7032}.
\newblock


\bibitem[Chen et~al\mbox{.}(2020)]%
        {chen2020}
\bibfield{author}{\bibinfo{person}{Ting Chen}, \bibinfo{person}{Simon Kornblith}, \bibinfo{person}{Mohammad Norouzi}, {and} \bibinfo{person}{Geoffrey Hinton}.} \bibinfo{year}{2020}\natexlab{}.
\newblock \showarticletitle{A Simple Framework for Contrastive Learning of Visual Representations}. In \bibinfo{booktitle}{\emph{ICML}}.
\newblock


\bibitem[Fan et~al\mbox{.}(2022)]%
        {Fan2022}
\bibfield{author}{\bibinfo{person}{Wenqi Fan}, \bibinfo{person}{Yao Ma}, \bibinfo{person}{Qing Li}, \bibinfo{person}{Jianping Wang}, \bibinfo{person}{Guoyong Cai}, \bibinfo{person}{Jiliang Tang}, {and} \bibinfo{person}{Dawei Yin}.} \bibinfo{year}{2022}\natexlab{}.
\newblock \showarticletitle{A Graph Neural Network Framework for Social Recommendations}.
\newblock \bibinfo{journal}{\emph{IEEE Transactions on Knowledge and Data Engineering}} \bibinfo{volume}{34}, \bibinfo{number}{5} (\bibinfo{year}{2022}), \bibinfo{pages}{2033--2047}.
\newblock


\bibitem[Gasteiger et~al\mbox{.}(2019)]%
        {gasteiger2018combining}
\bibfield{author}{\bibinfo{person}{Johannes Gasteiger}, \bibinfo{person}{Aleksandar Bojchevski}, {and} \bibinfo{person}{Stephan Günnemann}.} \bibinfo{year}{2019}\natexlab{}.
\newblock \showarticletitle{Combining Neural Networks with Personalized PageRank for Classification on Graphs}. In \bibinfo{booktitle}{\emph{ICLR}}.
\newblock


\bibitem[Giles et~al\mbox{.}(1998)]%
        {giles1998citeseer}
\bibfield{author}{\bibinfo{person}{C~Lee Giles}, \bibinfo{person}{Kurt~D Bollacker}, {and} \bibinfo{person}{Steve Lawrence}.} \bibinfo{year}{1998}\natexlab{}.
\newblock \showarticletitle{CiteSeer: An automatic citation indexing system}. In \bibinfo{booktitle}{\emph{Proceedings of the third ACM conference on Digital libraries}}. \bibinfo{pages}{89--98}.
\newblock


\bibitem[Gilmer et~al\mbox{.}(2017)]%
        {pmlr-v70-gilmer17a}
\bibfield{author}{\bibinfo{person}{Justin Gilmer}, \bibinfo{person}{Samuel~S. Schoenholz}, \bibinfo{person}{Patrick~F. Riley}, \bibinfo{person}{Oriol Vinyals}, {and} \bibinfo{person}{George~E. Dahl}.} \bibinfo{year}{2017}\natexlab{}.
\newblock \showarticletitle{Neural Message Passing for Quantum Chemistry}. In \bibinfo{booktitle}{\emph{ICML}}. \bibinfo{pages}{1263--1272}.
\newblock


\bibitem[Guo et~al\mbox{.}(2022)]%
        {guo2022}
\bibfield{author}{\bibinfo{person}{Zhihao Guo}, \bibinfo{person}{Feng Wang}, \bibinfo{person}{Kaixuan Yao}, \bibinfo{person}{Jiye Liang}, {and} \bibinfo{person}{Zhiqiang Wang}.} \bibinfo{year}{2022}\natexlab{}.
\newblock \showarticletitle{Multi-Scale Variational Graph AutoEncoder for Link Prediction}. In \bibinfo{booktitle}{\emph{WSDM}}.
\newblock


\bibitem[Haj~Ali and H{\"u}tt(2022)]%
        {haj2022inferring}
\bibfield{author}{\bibinfo{person}{Selim Haj~Ali} {and} \bibinfo{person}{Marc-Thorsten H{\"u}tt}.} \bibinfo{year}{2022}\natexlab{}.
\newblock \showarticletitle{Inferring missing edges in a graph from observed collective patterns}.
\newblock \bibinfo{journal}{\emph{Physical Review E}} \bibinfo{volume}{105}, \bibinfo{number}{6} (\bibinfo{year}{2022}), \bibinfo{pages}{064610}.
\newblock


\bibitem[Huang et~al\mbox{.}(2023b)]%
        {huang2023towards}
\bibfield{author}{\bibinfo{person}{Weiran Huang}, \bibinfo{person}{Mingyang Yi}, \bibinfo{person}{Xuyang Zhao}, {and} \bibinfo{person}{Zihao Jiang}.} \bibinfo{year}{2023}\natexlab{b}.
\newblock \showarticletitle{Towards the Generalization of Contrastive Self-Supervised Learning}. In \bibinfo{booktitle}{\emph{ICLR}}.
\newblock


\bibitem[Huang et~al\mbox{.}(2023a)]%
        {huang2023linkpredictiongraphneural}
\bibfield{author}{\bibinfo{person}{Zexi Huang}, \bibinfo{person}{Mert Kosan}, \bibinfo{person}{Arlei Silva}, {and} \bibinfo{person}{Ambuj Singh}.} \bibinfo{year}{2023}\natexlab{a}.
\newblock \showarticletitle{Link prediction without graph neural networks}.
\newblock \bibinfo{journal}{\emph{arXiv preprint arXiv:2305.13656}} (\bibinfo{year}{2023}).
\newblock


\bibitem[Kim et~al\mbox{.}(2025)]%
        {kim2025accurate}
\bibfield{author}{\bibinfo{person}{Junghun Kim}, \bibinfo{person}{Ka~Hyun Park}, \bibinfo{person}{Hoyoung Yoon}, {and} \bibinfo{person}{U Kang}.} \bibinfo{year}{2025}\natexlab{}.
\newblock \showarticletitle{Accurate Link Prediction for Edge-Incomplete Graphs via PU Learning}. In \bibinfo{booktitle}{\emph{Proceedings of the AAAI Conference on Artificial Intelligence}}, Vol.~\bibinfo{volume}{39}. \bibinfo{pages}{17877--17885}.
\newblock


\bibitem[Kingma and Ba(2017)]%
        {kingma2017}
\bibfield{author}{\bibinfo{person}{Diederik~P. Kingma} {and} \bibinfo{person}{Jimmy Ba}.} \bibinfo{year}{2017}\natexlab{}.
\newblock \showarticletitle{Adam: A Method for Stochastic Optimization}.
\newblock \bibinfo{journal}{\emph{arXiv preprint arXiv:1412.6980}} (\bibinfo{year}{2017}).
\newblock


\bibitem[Kipf and Welling(2016)]%
        {kipf2016}
\bibfield{author}{\bibinfo{person}{Thomas~N Kipf} {and} \bibinfo{person}{Max Welling}.} \bibinfo{year}{2016}\natexlab{}.
\newblock \showarticletitle{Variational graph auto-encoders}.
\newblock \bibinfo{journal}{\emph{arXiv preprint arXiv:1611.07308}} (\bibinfo{year}{2016}).
\newblock


\bibitem[Kossinets(2006)]%
        {kossinets2006effects}
\bibfield{author}{\bibinfo{person}{Gueorgi Kossinets}.} \bibinfo{year}{2006}\natexlab{}.
\newblock \showarticletitle{Effects of missing data in social networks}.
\newblock \bibinfo{journal}{\emph{Social networks}} \bibinfo{volume}{28}, \bibinfo{number}{3} (\bibinfo{year}{2006}), \bibinfo{pages}{247--268}.
\newblock


\bibitem[Kundu and Aakur(2023)]%
        {Kundu_2023_CVPR}
\bibfield{author}{\bibinfo{person}{Sanjoy Kundu} {and} \bibinfo{person}{Sathyanarayanan~N. Aakur}.} \bibinfo{year}{2023}\natexlab{}.
\newblock \showarticletitle{{IS-GGT}: Iterative Scene Graph Generation With Generative Transformers}. In \bibinfo{booktitle}{\emph{CVPR}}. \bibinfo{pages}{6292--6301}.
\newblock


\bibitem[Li et~al\mbox{.}(2024a)]%
        {Li2023evaluate}
\bibfield{author}{\bibinfo{person}{Juanhui Li}, \bibinfo{person}{Harry Shomer}, \bibinfo{person}{Haitao Mao}, \bibinfo{person}{Shenglai Zeng}, \bibinfo{person}{Yao Ma}, \bibinfo{person}{Neil Shah}, \bibinfo{person}{Jiliang Tang}, {and} \bibinfo{person}{Dawei Yin}.} \bibinfo{year}{2024}\natexlab{a}.
\newblock \showarticletitle{Evaluating graph neural networks for link prediction: current pitfalls and new benchmarking}. In \bibinfo{booktitle}{\emph{NeurIPS}}.
\newblock


\bibitem[Li et~al\mbox{.}(2024b)]%
        {Li2024toward}
\bibfield{author}{\bibinfo{person}{Wen-Zhi Li}, \bibinfo{person}{Chang-Dong Wang}, \bibinfo{person}{Jian-Huang Lai}, {and} \bibinfo{person}{Philip~S. Yu}.} \bibinfo{year}{2024}\natexlab{b}.
\newblock \showarticletitle{Towards Effective and Robust Graph Contrastive Learning With Graph Autoencoding}.
\newblock \bibinfo{journal}{\emph{IEEE Transactions on Knowledge and Data Engineering}} \bibinfo{volume}{36}, \bibinfo{number}{2} (\bibinfo{year}{2024}), \bibinfo{pages}{868--881}.
\newblock
\urldef\tempurl%
\url{https://doi.org/10.1109/TKDE.2023.3288280}
\showDOI{\tempurl}


\bibitem[Liang et~al\mbox{.}(2023)]%
        {liang2023knowledge}
\bibfield{author}{\bibinfo{person}{Ke Liang}, \bibinfo{person}{Yue Liu}, \bibinfo{person}{Sihang Zhou}, \bibinfo{person}{Wenxuan Tu}, \bibinfo{person}{Yi Wen}, \bibinfo{person}{Xihong Yang}, \bibinfo{person}{Xiangjun Dong}, {and} \bibinfo{person}{Xinwang Liu}.} \bibinfo{year}{2023}\natexlab{}.
\newblock \showarticletitle{Knowledge graph contrastive learning based on relation-symmetrical structure}.
\newblock \bibinfo{journal}{\emph{IEEE Transactions on Knowledge and Data Engineering}} \bibinfo{volume}{36}, \bibinfo{number}{1} (\bibinfo{year}{2023}), \bibinfo{pages}{226--238}.
\newblock


\bibitem[Ma et~al\mbox{.}(2024)]%
        {ma2024mixture}
\bibfield{author}{\bibinfo{person}{Li Ma}, \bibinfo{person}{Haoyu Han}, \bibinfo{person}{Juanhui Li}, \bibinfo{person}{Harry Shomer}, \bibinfo{person}{Hui Liu}, \bibinfo{person}{Xiaofeng Gao}, {and} \bibinfo{person}{Jiliang Tang}.} \bibinfo{year}{2024}\natexlab{}.
\newblock \showarticletitle{Mixture of Link Predictors on Graphs}. In \bibinfo{booktitle}{\emph{The Thirty-eighth Annual Conference on Neural Information Processing Systems}}.
\newblock


\bibitem[McCallum et~al\mbox{.}(2000)]%
        {mccallum2000automating}
\bibfield{author}{\bibinfo{person}{Andrew~Kachites McCallum}, \bibinfo{person}{Kamal Nigam}, \bibinfo{person}{Jason Rennie}, {and} \bibinfo{person}{Kristie Seymore}.} \bibinfo{year}{2000}\natexlab{}.
\newblock \showarticletitle{Automating the construction of internet portals with machine learning}.
\newblock \bibinfo{journal}{\emph{Information Retrieval}}  \bibinfo{volume}{3} (\bibinfo{year}{2000}), \bibinfo{pages}{127--163}.
\newblock


\bibitem[Oord et~al\mbox{.}(2018)]%
        {oord2019}
\bibfield{author}{\bibinfo{person}{Aaron van~den Oord}, \bibinfo{person}{Yazhe Li}, {and} \bibinfo{person}{Oriol Vinyals}.} \bibinfo{year}{2018}\natexlab{}.
\newblock \showarticletitle{Representation learning with contrastive predictive coding}.
\newblock \bibinfo{journal}{\emph{arXiv preprint arXiv:1807.03748}} (\bibinfo{year}{2018}).
\newblock


\bibitem[Park et~al\mbox{.}(2025)]%
        {park2025cimage}
\bibfield{author}{\bibinfo{person}{Jongwon Park}, \bibinfo{person}{Heesoo Jung}, {and} \bibinfo{person}{Hogun Park}.} \bibinfo{year}{2025}\natexlab{}.
\newblock \showarticletitle{CIMAGE: Exploiting the Conditional Independence in Masked Graph Auto-encoders}. In \bibinfo{booktitle}{\emph{Proceedings of the Eighteenth ACM International Conference on Web Search and Data Mining}}. \bibinfo{pages}{10--19}.
\newblock


\bibitem[Sarkar et~al\mbox{.}(2011)]%
        {sarkar2011theoretical}
\bibfield{author}{\bibinfo{person}{Purnamrita Sarkar}, \bibinfo{person}{Deepayan Chakrabarti}, {and} \bibinfo{person}{Andrew~W Moore}.} \bibinfo{year}{2011}\natexlab{}.
\newblock \showarticletitle{Theoretical Justification of Popular Link Prediction Heuristics.}. In \bibinfo{booktitle}{\emph{IJCAI}}.
\newblock


\bibitem[Shchur et~al\mbox{.}(2018)]%
        {shchur2019pitfallsgraphneuralnetwork}
\bibfield{author}{\bibinfo{person}{Oleksandr Shchur}, \bibinfo{person}{Maximilian Mumme}, \bibinfo{person}{Aleksandar Bojchevski}, {and} \bibinfo{person}{Stephan G{\"u}nnemann}.} \bibinfo{year}{2018}\natexlab{}.
\newblock \showarticletitle{Pitfalls of graph neural network evaluation}.
\newblock \bibinfo{journal}{\emph{arXiv preprint arXiv:1811.05868}} (\bibinfo{year}{2018}).
\newblock


\bibitem[Shen et~al\mbox{.}(2023)]%
        {Shen2023}
\bibfield{author}{\bibinfo{person}{Xiao Shen}, \bibinfo{person}{Dewang Sun}, \bibinfo{person}{Shirui Pan}, \bibinfo{person}{Xi Zhou}, {and} \bibinfo{person}{Laurence~T Yang}.} \bibinfo{year}{2023}\natexlab{}.
\newblock \showarticletitle{Neighbor contrastive learning on learnable graph augmentation}. In \bibinfo{booktitle}{\emph{AAAI}}. \bibinfo{pages}{9782--9791}.
\newblock


\bibitem[Shiao et~al\mbox{.}(2023)]%
        {shiao2023link}
\bibfield{author}{\bibinfo{person}{William Shiao}, \bibinfo{person}{Zhichun Guo}, \bibinfo{person}{Tong Zhao}, \bibinfo{person}{Evangelos~E. Papalexakis}, \bibinfo{person}{Yozen Liu}, {and} \bibinfo{person}{Neil Shah}.} \bibinfo{year}{2023}\natexlab{}.
\newblock \showarticletitle{Link Prediction with Non-Contrastive Learning}. In \bibinfo{booktitle}{\emph{ICLR}}.
\newblock


\bibitem[Tan et~al\mbox{.}(2022)]%
        {tan2022}
\bibfield{author}{\bibinfo{person}{Qiaoyu Tan}, \bibinfo{person}{Ninghao Liu}, \bibinfo{person}{Xiao Huang}, \bibinfo{person}{Soo-Hyun Choi}, \bibinfo{person}{Li Li}, \bibinfo{person}{Rui Chen}, {and} \bibinfo{person}{Xia Hu}.} \bibinfo{year}{2022}\natexlab{}.
\newblock \showarticletitle{{S2GAE}: Self-Supervised Graph Autoencoders are Generalizable Learners with Graph Masking}. In \bibinfo{booktitle}{\emph{WSDM}}.
\newblock


\bibitem[Tan et~al\mbox{.}(2024)]%
        {tan2024contrastive}
\bibfield{author}{\bibinfo{person}{Zhiquan Tan}, \bibinfo{person}{Yifan Zhang}, \bibinfo{person}{Jingqin Yang}, {and} \bibinfo{person}{Yang Yuan}.} \bibinfo{year}{2024}\natexlab{}.
\newblock \showarticletitle{Contrastive Learning is Spectral Clustering on Similarity Graph}. In \bibinfo{booktitle}{\emph{ICLR}}.
\newblock


\bibitem[Tian(2022)]%
        {NEURIPS2022_Tian}
\bibfield{author}{\bibinfo{person}{Yuandong Tian}.} \bibinfo{year}{2022}\natexlab{}.
\newblock \showarticletitle{Understanding Deep Contrastive Learning via Coordinate-wise Optimization}. In \bibinfo{booktitle}{\emph{NeurIPS}}, \bibfield{editor}{\bibinfo{person}{S.~Koyejo}, \bibinfo{person}{S.~Mohamed}, \bibinfo{person}{A.~Agarwal}, \bibinfo{person}{D.~Belgrave}, \bibinfo{person}{K.~Cho}, {and} \bibinfo{person}{A.~Oh}} (Eds.). \bibinfo{pages}{19511--19522}.
\newblock


\bibitem[Tian(2023)]%
        {tian2023understanding}
\bibfield{author}{\bibinfo{person}{Yuandong Tian}.} \bibinfo{year}{2023}\natexlab{}.
\newblock \showarticletitle{Understanding the Role of Nonlinearity in Training Dynamics of Contrastive Learning}. In \bibinfo{booktitle}{\emph{ICLR}}.
\newblock


\bibitem[Wang et~al\mbox{.}(2022)]%
        {Wang2022}
\bibfield{author}{\bibinfo{person}{Ruijia Wang}, \bibinfo{person}{Xiao Wang}, \bibinfo{person}{Chuan Shi}, {and} \bibinfo{person}{Le Song}.} \bibinfo{year}{2022}\natexlab{}.
\newblock \showarticletitle{Uncovering the structural fairness in graph contrastive learning}. In \bibinfo{booktitle}{\emph{NeurIPS}}. \bibinfo{pages}{32465--32473}.
\newblock


\bibitem[Wang and Isola(2020)]%
        {pmlr-v119-wang20k}
\bibfield{author}{\bibinfo{person}{Tongzhou Wang} {and} \bibinfo{person}{Phillip Isola}.} \bibinfo{year}{2020}\natexlab{}.
\newblock \showarticletitle{Understanding Contrastive Representation Learning through Alignment and Uniformity on the Hypersphere}. In \bibinfo{booktitle}{\emph{ICML}}. \bibinfo{pages}{9929--9939}.
\newblock


\bibitem[Wang et~al\mbox{.}(2024)]%
        {wang2024}
\bibfield{author}{\bibinfo{person}{Xiyuan Wang}, \bibinfo{person}{Haotong Yang}, {and} \bibinfo{person}{Muhan Zhang}.} \bibinfo{year}{2024}\natexlab{}.
\newblock \showarticletitle{Neural Common Neighbor with Completion for Link Prediction}. In \bibinfo{booktitle}{\emph{ICLR}}.
\newblock


\bibitem[Wang et~al\mbox{.}(2023)]%
        {Wang2023recommend}
\bibfield{author}{\bibinfo{person}{Yu Wang}, \bibinfo{person}{Yuying Zhao}, \bibinfo{person}{Yi Zhang}, {and} \bibinfo{person}{Tyler Derr}.} \bibinfo{year}{2023}\natexlab{}.
\newblock \showarticletitle{Collaboration-Aware Graph Convolutional Network for Recommender Systems}. In \bibinfo{booktitle}{\emph{the Web Conference}}. \bibinfo{pages}{91–101}.
\newblock


\bibitem[Xiao et~al\mbox{.}(2023)]%
        {xiao2023simple}
\bibfield{author}{\bibinfo{person}{Teng Xiao}, \bibinfo{person}{Huaisheng Zhu}, \bibinfo{person}{Zhengyu Chen}, {and} \bibinfo{person}{Suhang Wang}.} \bibinfo{year}{2023}\natexlab{}.
\newblock \showarticletitle{Simple and Asymmetric Graph Contrastive Learning without Augmentations}. In \bibinfo{booktitle}{\emph{Thirty-seventh Conference on Neural Information Processing Systems}}.
\newblock


\bibitem[Yang et~al\mbox{.}(2023)]%
        {yang2023}
\bibfield{author}{\bibinfo{person}{Hao-Wei Yang}, \bibinfo{person}{Ming-Yi Chang}, {and} \bibinfo{person}{Chih-Ya Shen}.} \bibinfo{year}{2023}\natexlab{}.
\newblock \showarticletitle{Enhancing Link Prediction with Self-Discriminating Augmentation for Structure-Aware Contrastive Learning}. In \bibinfo{booktitle}{\emph{ECAI}}. \bibinfo{pages}{2842--2849}.
\newblock


\bibitem[Yang et~al\mbox{.}(2015)]%
        {Yang2015}
\bibfield{author}{\bibinfo{person}{Yang Yang}, \bibinfo{person}{Ryan~N. Lichtenwalter}, {and} \bibinfo{person}{Nitesh~V. Chawla}.} \bibinfo{year}{2015}\natexlab{}.
\newblock \showarticletitle{Evaluating link prediction methods}.
\newblock \bibinfo{journal}{\emph{Knowledge and Information Systems}} \bibinfo{volume}{45}, \bibinfo{number}{3} (\bibinfo{year}{2015}), \bibinfo{pages}{751--782}.
\newblock


\bibitem[Yin et~al\mbox{.}(2022)]%
        {yin2022}
\bibfield{author}{\bibinfo{person}{Yihang Yin}, \bibinfo{person}{Qingzhong Wang}, \bibinfo{person}{Siyu Huang}, \bibinfo{person}{Haoyi Xiong}, {and} \bibinfo{person}{Xiang Zhang}.} \bibinfo{year}{2022}\natexlab{}.
\newblock \showarticletitle{{AutoGCL}: Automated Graph Contrastive Learning via Learnable View Generators}. In \bibinfo{booktitle}{\emph{AAAI}}.
\newblock


\bibitem[You et~al\mbox{.}(2020)]%
        {you2020}
\bibfield{author}{\bibinfo{person}{Yuning You}, \bibinfo{person}{Tianlong Chen}, \bibinfo{person}{Yongduo Sui}, \bibinfo{person}{Ting Chen}, \bibinfo{person}{Zhangyang Wang}, {and} \bibinfo{person}{Yang Shen}.} \bibinfo{year}{2020}\natexlab{}.
\newblock \showarticletitle{Graph Contrastive Learning with Augmentations}. In \bibinfo{booktitle}{\emph{NeurIPS}}.
\newblock


\bibitem[Yun et~al\mbox{.}(2022)]%
        {yun2022}
\bibfield{author}{\bibinfo{person}{Seongjun Yun}, \bibinfo{person}{Seoyoon Kim}, \bibinfo{person}{Junhyun Lee}, \bibinfo{person}{Jaewoo Kang}, {and} \bibinfo{person}{Hyunwoo~J. Kim}.} \bibinfo{year}{2022}\natexlab{}.
\newblock \showarticletitle{{Neo-GNNs}: Neighborhood Overlap-aware Graph Neural Networks for Link Prediction}. In \bibinfo{booktitle}{\emph{NeurIPS}}.
\newblock


\bibitem[Zangari et~al\mbox{.}(2024)]%
        {zangari2024link}
\bibfield{author}{\bibinfo{person}{Lorenzo Zangari}, \bibinfo{person}{Domenico Mandaglio}, {and} \bibinfo{person}{Andrea Tagarelli}.} \bibinfo{year}{2024}\natexlab{}.
\newblock \showarticletitle{Link prediction on multilayer networks through learning of within-layer and across-layer node-pair structural features and node embedding similarity}. In \bibinfo{booktitle}{\emph{Proceedings of the ACM Web Conference 2024}}. \bibinfo{pages}{924--935}.
\newblock


\bibitem[Zhang and Chen(2018)]%
        {zhang2018}
\bibfield{author}{\bibinfo{person}{Muhan Zhang} {and} \bibinfo{person}{Yixin Chen}.} \bibinfo{year}{2018}\natexlab{}.
\newblock \showarticletitle{Link Prediction Based on Graph Neural Networks}. In \bibinfo{booktitle}{\emph{NeurIPS}}.
\newblock


\bibitem[Zhang et~al\mbox{.}(2023)]%
        {zhang2023line}
\bibfield{author}{\bibinfo{person}{Zehua Zhang}, \bibinfo{person}{Shilin Sun}, \bibinfo{person}{Guixiang Ma}, {and} \bibinfo{person}{Caiming Zhong}.} \bibinfo{year}{2023}\natexlab{}.
\newblock \showarticletitle{Line graph contrastive learning for link prediction}.
\newblock \bibinfo{journal}{\emph{Pattern Recognition}}  \bibinfo{volume}{140} (\bibinfo{year}{2023}), \bibinfo{pages}{109537}.
\newblock


\bibitem[Zhao et~al\mbox{.}(2022)]%
        {zhao2022}
\bibfield{author}{\bibinfo{person}{Tong Zhao}, \bibinfo{person}{Gang Liu}, \bibinfo{person}{Daheng Wang}, \bibinfo{person}{Wenhao Yu}, {and} \bibinfo{person}{Meng Jiang}.} \bibinfo{year}{2022}\natexlab{}.
\newblock \showarticletitle{Learning from Counterfactual Links for Link Prediction}. In \bibinfo{booktitle}{\emph{ICML}}. \bibinfo{pages}{26911--26926}.
\newblock


\bibitem[Zhou et~al\mbox{.}(2023)]%
        {zhou2023combating}
\bibfield{author}{\bibinfo{person}{Zhanke Zhou}, \bibinfo{person}{Jiangchao Yao}, \bibinfo{person}{Jiaxu Liu}, \bibinfo{person}{Xiawei Guo}, \bibinfo{person}{quanming yao}, \bibinfo{person}{LI He}, \bibinfo{person}{Liang Wang}, \bibinfo{person}{Bo Zheng}, {and} \bibinfo{person}{Bo Han}.} \bibinfo{year}{2023}\natexlab{}.
\newblock \showarticletitle{Combating Bilateral Edge Noise for Robust Link Prediction}. In \bibinfo{booktitle}{\emph{Thirty-seventh Conference on Neural Information Processing Systems}}.
\newblock


\bibitem[Zhu et~al\mbox{.}(2020)]%
        {zhu2020}
\bibfield{author}{\bibinfo{person}{Yanqiao Zhu}, \bibinfo{person}{Yichen Xu}, \bibinfo{person}{Feng Yu}, \bibinfo{person}{Qiang Liu}, \bibinfo{person}{Shu Wu}, {and} \bibinfo{person}{Liang Wang}.} \bibinfo{year}{2020}\natexlab{}.
\newblock \showarticletitle{Deep graph contrastive representation learning}.
\newblock \bibinfo{journal}{\emph{arXiv preprint arXiv:2006.04131}} (\bibinfo{year}{2020}).
\newblock


\bibitem[Zhu et~al\mbox{.}(2021)]%
        {zhu2021}
\bibfield{author}{\bibinfo{person}{Yanqiao Zhu}, \bibinfo{person}{Yichen Xu}, \bibinfo{person}{Feng Yu}, \bibinfo{person}{Qiang Liu}, \bibinfo{person}{Shu Wu}, {and} \bibinfo{person}{Liang Wang}.} \bibinfo{year}{2021}\natexlab{}.
\newblock \showarticletitle{Graph Contrastive Learning with Adaptive Augmentation}. In \bibinfo{booktitle}{\emph{the Web Conference}}.
\newblock


\end{thebibliography}


\end{CJK}
\end{document}